%% file: challenge_eval.tex
\documentclass[11pt,letterpaper]{article}
\usepackage[hyperref]{emnlp2017-pagenum}
\usepackage{times}
\usepackage{latexsym}
\usepackage[utf8x]{inputenc}
\usepackage{pifont}
\usepackage{comment}
\newcommand{\cmark}{\ding{51}}%
\newcommand{\xmark}{\ding{55}}%
\usepackage{url}

\emnlpfinalcopy

\def\emnlppaperid{477}

\title{A Challenge Set Approach to Evaluating Machine Translation}

\author{Pierre Isabelle and Colin Cherry  \\
  National Research Council Canada\\
  {\tt first.last@nrc-cnrc.gc.ca} \\\And
  George Foster \\
  Google\thanks{~~Work performed while at NRC.}  \\
  {\tt fosterg@google.com} \\}

\date{}

\begin{document}
\maketitle
\begin{abstract}
  Neural machine translation represents an exciting leap forward in translation 	
  quality. 
    But what longstanding weaknesses
    does it resolve, and which remain?
    We address these questions with a challenge set approach to translation evaluation and 
    error analysis.
    A challenge set consists of 
    a small set of sentences,
    each hand-designed to probe a system's capacity to bridge a particular 
    structural divergence between languages. 
    To exemplify this approach, we present an English--French challenge set, and
    use it to analyze
    phrase-based and neural systems.
    The resulting analysis provides not only a more fine-grained picture of the
    strengths of neural systems, but also insight into which linguistic phenomena remain out of 
    reach.
\end{abstract}

\section{Introduction}



The advent of neural techniques in machine translation (MT)
\cite{Kalchbrenner13,Cho14,Sutskever14} has led to profound improvements in MT
quality. For ``easy'' language pairs such as English/French or English/Spanish
in particular, neural (NMT) systems are much closer to human performance than
previous statistical techniques~\cite{Wu16}. This puts pressure on automatic
evaluation metrics such as BLEU~\cite{Papineni:ACL2002}, which exploit
surface-matching heuristics that are relatively insensitive to subtle
differences.  As NMT continues to improve, these metrics will
inevitably lose their effectiveness. Another challenge posed by NMT systems is
their opacity: while it was usually clear which phenomena were ill-handled by
previous statistical systems---and why---these questions are more difficult to
answer for NMT.

\begin{figure}[t]
\begin{tabular}{|lp{0.8\columnwidth}|}
\hline
Src & The repeated calls from his mother \textbf{should} have alerted us. \\
Ref & Les appels répétés de sa mère \textbf{auraient} dû nous alerter. \\
Sys & Les appels répétés de sa mère devraient nous avoir alertés.\\
\hline
\multicolumn{2}{|l|}{Is the subject-verb agreement correct (y/n)? \textbf{Yes}}\\
\hline
\end{tabular}
\caption{Example challenge set question.\label{fig:example}}
\vspace{-5pt}
\end{figure}

We propose a new evaluation methodology centered around
a \textit{challenge set} of difficult examples that are designed using expert linguistic
knowledge to probe an MT system's capabilities. This methodology is
complementary to the standard practice of randomly selecting a test set from
``real text,'' which remains necessary in order to predict
performance on new text. By concentrating on difficult examples, a challenge
set is intended to provide a stronger signal to developers. 
Although we believe
that the general approach is compatible with automatic metrics, we used manual
evaluation for the work presented here. 
Our challenge set consists of short sentences that each focus
on one particular phenomenon, which makes it easy to collect reliable manual assessments of MT output
by asking direct yes-no questions.
An example is shown in Figure~\ref{fig:example}.

We generated a challenge set for English to French translation by canvassing
areas of linguistic divergence between the two language pairs, especially those
where errors would be made visible by French morphology. 
Example choice was also partly motivated by
extensive knowledge of the weaknesses of phrase-based MT (PBMT). Neither of
these characteristics is essential to our method, however, which we envisage
evolving as NMT progresses. We used our challenge set to evaluate
in-house PBMT and NMT systems as well as Google's GNMT system.

In addition to proposing the novel idea of a challenge set evaluation, our
contribution includes our annotated English--French challenge set, which we
provide in both formatted text and machine-readable formats (see supplemental
materials). We also supply
further evidence that NMT is systematically better than PBMT, even when BLEU
score differences are small. Finally, we give an analysis of the challenges
that remain to be solved in NMT, an area that has received little attention thus far.

\section{Related Work}
A number of recent papers have evaluated NMT using broad performance metrics.
The WMT 2016 News Translation Task~\cite{Bojar:WMT2016} evaluated submitted systems according to both BLEU and human judgments.
NMT systems were submitted to 9 of the 12 translation directions, winning 4 of these and tying for first or second in the other 5, according to the official human ranking.
Since then, controlled comparisons have used BLEU to show that NMT outperforms strong PBMT systems on 30 translation directions from the United Nations Parallel Corpus~\cite{Junczys-Dowmunt:IWSLT2016}, 
and on the IWSLT English-Arabic tasks~\cite{Durrani:IWSLT2016}.
These evaluations indicate that NMT performs better on average than previous technologies, but they do not help us understand what aspects of the translation have improved.

Some groups have conducted more detailed error analyses.
\newcite{Bentivogli:EMNLP2016} carried out a number of experiments on IWSLT 2015 English-German evaluation data,
where they compare machine outputs to professional post-edits in order to automatically detect a number of error categories.
Compared to PBMT, NMT required less post-editing effort overall, with substantial improvements in lexical, morphological and word order errors.
NMT consistently out-performed PBMT, but its performance degraded faster as sentence length increased.
Later, \newcite{Toral:EACL2017} conducted a similar study, examining the outputs of competition-grade systems for the 9 WMT 2016 directions that included NMT competitors.
They reached similar conclusions regarding morphological inflection and word order, but found an even greater degradation in NMT performance as sentence length increased,
perhaps due to these systems' use of subword units.

Most recently, \newcite{Sennrich:arXiv2016} proposed an approach to perform targeted evaluations of NMT
through the use of contrastive translation pairs.
This method introduces a particular type of error automatically in reference sentences, and then checks whether the NMT system's
conditional probability model prefers the original reference or the corrupted version.
Using this technique, they are able to determine that a recently-proposed character-based model improves generalization on unseen words,
but at the cost of introducing new grammatical errors.

Our approach differs from these studies in a number of ways.
First, whereas others have analyzed sentences drawn from an existing bitext,
we conduct our study on sentences that are manually constructed to exhibit canonical examples of specific linguistic phenomena. 
We focus on phenomena that we expect to be more difficult than average,
resulting in a particularly challenging MT test suite~\cite{King90}.
These sentences are designed to dive deep into linguistic phenomena of
interest, and to provide a much finer-grained analysis of the strengths and weaknesses of existing technologies, including NMT systems.

However, this strategy also necessitates that we work on fewer sentences.
We leverage the small size of our challenge set to manually evaluate whether the system's actual output correctly handles our phenomena of interest.
Manual evaluation side-steps some of the pitfalls that can come with \newcite{Sennrich:arXiv2016}'s contrastive pairs,
as a ranking of two contrastive sentences may not necessarily reflect whether the error in question will occur in the system's actual output.

\section{Challenge Set Evaluation}

Our challenge set is meant to measure the ability of MT systems to
deal with some of the more difficult problems that arise in
translating English into French. This particular language pair happened to be most
convenient for us, but similar sets can be built for any
language pair.

One aspect of MT performance
excluded from
our evaluation is robustness to sparse data. To control for this, when crafting
source and reference sentences, we chose words
that occurred
at least 100 times in our training corpus
(section~\ref{ssec:data}).\footnote{With two exceptions: \emph{spilt} (58
  occurrences), which is part of an idiomatic phrase, 
and \emph{guitared} (0 occurrences), which is meant to test the ability to deal with "nonce words" as discussed in section 5.}

The challenging aspect of the test set we are presenting stems from the fact
that the source English sentences have been chosen so that their
closest French equivalent will be \textit{structurally divergent} from the
source in some crucial way.  Translational divergences have been extensively
studied in the past---see for example \cite{Vinay:58, Dorr:94}. We expect the
level of difficulty of an MT test set to correlate well with its density in
divergence phenomena, which we
classify into three main types: morpho-syntactic, lexico-syntactic 
and purely syntactic divergences.

\subsection{Morpho-syntactic divergences}

In some languages, word morphology (e.g. inflections) carries more grammatical information than in others.
When translating a word towards the richer language, there is a need
 to recover additional grammatically-relevant information from the context of the target
 language word. Note that we only include in our set cases where the relevant information is available in the {\em linguistic} context.\footnote{The so-called Winograd Schema Challenges (en.wikipedia.org/wiki/Winograd\_Schema\_Challenge)
often involve cases where common-sense reasoning is required to correctly
choose between two potential antecedent phrases for a pronoun. Such cases become 
En $\rightarrow$ Fr translation challenges if the relevant English pronoun is {\em they} and its
alternative antecedents happen to have different grammatical genders in French: {\em
  they} $\rightarrow$ {\em ils/elles}.}

One particularly important case of morpho-syntactic divergence is that of \emph{subject--verb agreement}. French verbs typically have more than 30 different inflected forms, while English 
verbs typically have 4 or 5. As a result, English verb forms strongly underspecify their French counterparts. Much of the missing information must be filled in through forced agreement in person, number and gender with the grammatical subject of the 
verb. But extracting these parameters can prove difficult. For example, the agreement features of a coordinated noun phrase are a complex function of the coordinated elements: a) the gender is feminine if all conjuncts are feminine, otherwise masculine wins; b) the conjunct with the smallest person (p1$<$p2$<$p3) wins; and c) the number is always plural when the coordination is ``et'' (``and'') but the case is more complex with ``ou'' (``or''). 

A second example of morpho-syntactic divergence between English and French is the more explicit marking of the {\em subjunctive mood} in French subordinate clauses. In the following example, the verb ``partiez'', unlike its English counterpart, is marked as subjunctive:
\begin{quote}
He demanded that you leave immediately. $\rightarrow$ Il a exigé que vous
  \emph{partiez} immédiatement.
\end{quote}
When translating an English verb within a subordinate clause, the context must be examined for 
possible subjunctive triggers. Typically these are specific lexical items found in a governing position with respect to the subordinate clause: verbs such as ``exiger que'', adjectives such as ``regrettable que'' 
or subordinate conjunctions such as ``à condition que''. 

\subsection{Lexico-syntactic divergences}

Syntactically governing words such as verbs tend to impose specific requirements on 
their complements: they {\em subcategorize} for complements of a certain syntactic type. 
But a source language governor and its target language counterpart can diverge 
on their respective requirements. The translation of such words must then trigger adjustments 
in the target language complement pattern. We can only examine here a few of the types 
instantiated in our challenge set.

A good example is \emph{argument switching}. This refers to the situation where 
the translation of a source verb V$_{s}$ as V$_{t}$ is correct 
but only provided the arguments (usually the subject and the object) are flipped around. 
The translation of ``to miss'' as ``manquer à'' is such a case:
\begin{quote}
John misses Mary  $\rightarrow$ Mary \emph{manque à} John.
\end{quote}  
Failing to perform the switch results in a severe case of mistranslation.

A second example of lexico-syntactic divergence is that of ``crossing movement'' verbs.
Consider the following example:
\begin{quote}
  Terry swam across the river $\rightarrow$ Terry \emph{a traversé} la rivière \emph{à la
  nage}.
\end{quote}
The French translation could be glossed as, ``Terry crossed the river by swimming.''
A literal translation such as ``Terry a nagé à travers la rivière,'' is ruled out.

\subsection{Syntactic divergences}

Some syntactic divergences are not relative to the presence of a particular lexical 
item but rather stem from differences in the set of available syntactic patterns. 
Source-language instances of structures missing from the target language 
must be mapped onto equivalent structures. Here are some of the types appearing in 
our challenge set.

The position of French pronouns is a major case of divergence from English. 
French is basically an SVO language like English but it departs from that 
canonical order when post-verbal complements are pronominalized: the pronouns must then be rendered as
\emph{proclitics}, that is phonetically attached to the verb on its left side.

\begin{quote}
He gave Mary a book.  $\rightarrow$ Il a donné un livre à Marie.

He gave$_{i}$ it$_{j}$ to her$_{k}$. $\rightarrow$  Il \emph{le$_{j}$
  lui$_{k}$} a donné$_{i}$.
\end{quote}

Another example of syntactic divergence between English and
French is that of \emph{stranded prepositions}. In both languages, an operation 
known as ``WH-movement'' will move a relativized or questioned element to the 
front of the clause containing it. When this element happens to be a prepositional
phrase, English offers the option to leave the preposition in its normal place, 
fronting only its pronominalized object. In French, the preposition is always 
fronted alongside its object:
\begin{quote}
The girl whom$_{i}$ he was dancing with$_{j}$ is rich. $\rightarrow$ La fille
\emph{avec$_{j}$ qui$_{i}$} il dansait est riche.
\end{quote}

A final example of syntactic divergence is the use of the so-called
\emph{middle voice}. While English uses the passive voice in
agentless generic statements, French tends to prefer the use of a
special pronominal construction where the pronoun ``se'' has no real
referent:
\begin{quote}
Caviar is eaten with bread. $\rightarrow$ Le caviar \emph{se mange} avec du pain.
\end{quote}

This completes our exemplification of morpho-syntactic, lexico-syntactic and
purely syntactic divergences.  Our actual test set includes
several more subcategories of each type.  The ability of MT systems to deal with each such subcategory is then tested using at least three different test sentences. We use short test
sentences so as to keep the targeted divergence in focus.
The 108
sentences that constitute our current challenge set
can be found in Appendix~\ref{appendix:challenge_set}.

\subsection{Evaluation Methodology} \label{ssec:evaluation-methodology}

Given the very small size of our challenge set, it is easy to perform a human
evaluation of the respective outputs of a handful of different systems. The
obvious advantage is that the assessment is then absolute instead of relative
to one or a few reference translations.

The intent of each challenge sentence is to test one and only one system
capability, namely that of coping correctly with the particular associated
divergence subtype.  As illustrated in Figure~\ref{fig:example}, we provide annotators with a question that specifies the
divergence phenomenon currently being tested, along with a reference
translation with the areas of divergence highlighted.  As a result, judgments
become straightforward: was the targeted divergence correctly bridged, yes or
no?\footnote{Sometimes the system produces a translation that circumvents the
  divergence issue. For example, it may dodge a divergence involving adverbs by
  reformulating the translation to use an adjective instead.  In these rare
  cases, we instruct our annotators to abstain from making a judgment,
  regardless of whether the translation is correct or not.  } There is no need
to mentally average over a number of different aspects of the test sentence as
one does when rating the global translation quality of a sentence, e.g. on a
5-point scale.  However, we acknowledge that measuring translation performance
on complex sentences exhibiting many different phenomena remains crucial.  We
see our approach as being complementary to evaluations of overall translation
quality.

One consequence of our divergence-focused approach is that faulty translations
will be judged as successes when the faults lie outside of the targeted
divergence zone. However, this problem is mitigated by our use of short test
sentences.

\section{Machine Translation Systems} \label{sec:systems}

We trained state-of-the-art neural and phrase-based systems for
English-French translation on data from the WMT 2014 evaluation.

\subsection{Data} \label{ssec:data}

We used the LIUM shared-task subset of the WMT 2014
corpora,\footnote{http://www.statmt.org/wmt14/translation-task.html\\http://www-lium.univ-lemans.fr/$\scriptstyle\sim$schwenk/nnmt-shared-task}
retaining the provided tokenization and corpus organization, but mapping
characters to lowercase. Table~\ref{tab:data} gives corpus statistics.

\begin{table}
  \begin{center}
  \begin{tabular}{lrrr}
    \hline
    corpus & lines & en words & fr words \\
    \hline
    train & 12.1M & 304M & 348M \\
    mono & 15.9M & ---- & 406M \\
    dev & 6003 & 138k & 155k \\
    test & 3003 & 71k & 81k \\
    \hline
  \end{tabular}
  \caption{Corpus statistics. The WMT12/13 eval sets are used for dev, and the WMT14 eval set
    is used for test.}
  \label{tab:data}
  \end{center}
\end{table}

\subsection{Phrase-based systems}

To ensure a competitive PBMT baseline, we performed 
phrase extraction using both IBM4 and HMM alignments 
with a phrase-length limit of 7; after frequency pruning,
the resulting phrase table
contained 516M entries. For each extracted phrase pair, we collected statistics
for the hierarchical reordering model of Galley and Manning
\shortcite{Galley08}.

We trained an NNJM model \cite{Devlin:14} on the HMM-aligned training corpus,
with input and output vocabulary sizes of 64k and 32k.
Words not in the vocabulary were mapped to one
of 100 mkcls classes. We trained for 60 epochs of 20k $\times$
128 minibatches, yielding a final dev-set perplexity of 6.88.

Our set of log-linear features consisted of forward and backward Kneser-Ney
smoothed phrase probabilities and HMM lexical probabilities (4 features); hierarchical
reordering probabilities (6); the NNJM probability (1); a set of sparse
features as described by \newcite{Cherry2013} (10,386); word-count and
distortion penalties (2); and 5-gram language models trained on the French
half of the training corpus and the French monolingual corpus (2).
Tuning was carried out using batch lattice MIRA
\cite{CherryFoster2012}.  Decoding used the cube-pruning algorithm of Huang and
Chiang \shortcite{HuangChiang2007}, with a distortion limit of 7.

We include two phrase-based systems in our comparison: PBMT-1 has data
conditions that exactly match those of the NMT system,
in that it does not use the language model trained on the French monolingual corpus, while
PBMT-2 uses both language models.

\subsection{Neural systems} \label{ssec:neural}

To build our NMT system, we used the Nematus
toolkit,\footnote{https://github.com/rsennrich/nematus} which implements a
single-layer neural sequence-to-sequence architecture with attention~\cite{Bahdanau15} and gated
recurrent units \cite{Cho14}. We used 512-dimensional word embeddings with source
and target vocabulary sizes of 90k, and 1024-dimensional state vectors.  The
model contains 172M parameters.

We preprocessed the data using a 
BPE model learned from
source and target corpora \cite{Sennrich15}. Sentences longer than 50 words
were discarded. Training used the Adadelta algorithm~\cite{Zeiler12}, with a
minibatch size of 100 and gradients clipped to 1.0. It ran for 5 epochs,
writing a checkpoint model every 30k minibatches. Following
\newcite{Junczys-Dowmunt16}, we averaged the parameters from the last 8
checkpoints.
To decode, we used the AmuNMT decoder \cite{Junczys-Dowmunt:IWSLT2016} with
a beam size of 4.

While our primary results will focus on the above PBMT and NMT systems, where we
can describe replicable configurations, we have also evaluated Google's production
system,\footnote{https://translate.google.com}
which has recently moved to NMT~\cite{Wu16}.
Notably, the ``GNMT'' system uses (at least) 8 encoder and 8 decoder layers, compared to our 1 layer for each,
and it is trained on corpora that are ``two to three decimal orders of magnitudes bigger than the WMT.''
The evaluated outputs were downloaded in December 2016.

\section{Experiments}

The 108-sentence English--French challenge set presented in Appendix~\ref{appendix:challenge_set} was
submitted to the four MT systems described in section~\ref{sec:systems}: PBMT-1, PBMT-2,
NMT, and GNMT.
Three bilingual native speakers of French rated each translated sentence as
either a success or a failure according to the protocol described in
section~\ref{ssec:evaluation-methodology}.
For example, the 26
sentences of the subcategories S1--S5 of Appendix~\ref{appendix:challenge_set}
are all about different cases of subject-verb agreement. The corresponding
translations were judged successful if and only if the translated verb
correctly agrees with the translated subject.

The different system outputs for each source sentence were grouped together to reduce
the burden on the annotators. That is, in figure~\ref{fig:example}, annotators
were asked to answer the question for each of four outputs, rather than just
one as shown. The outputs were listed in random order, without
identification. Questions were also presented in random order to each
annotator. Appendix~\ref{appendix:annotator_instructions} in the supplemental materials
contains the instructions shown to the annotators.

\subsection{Quantitative comparison}

\begin{table*}
  \begin{center}
  \begin{tabular}{lrrrrr}
    \hline
    Divergence type & PBMT-1 & PBMT-2 &  NMT & Google NMT & Agreement\\
    \hline
    Morpho-syntactic & 16\%	& 16\%  & 72\%  & 65\%   & 94\% \\
    Lexico-syntactic & 42\%	& 46\%  & 52\%	& 62\%   & 94\% \\
    Syntactic        & 33\%     & 33\%	& 40\%  & 75\%   & 81\% \\
    Overall          & 31\%     & 32\%  & 53\%  & 68\%   & 89\% \\
    \hline
    WMT BLEU& 34.2 & 36.5 & 36.9 & --- & --- \\
    \hline
  \end{tabular}
  \caption{Summary performance statistics for each system under study,
    including challenge set success rate grouped by linguistic category
    (aggregating all positive judgments and dividing by total judgments), as
    well as BLEU scores on the WMT 2014 test set. The final column gives the
    proportion of system outputs on which all three annotators agreed.}
  \label{tab:results}
  \end{center}
\end{table*}

Table~\ref{tab:results} summarizes our results in terms of percentage of successful translations,
globally and over each main type of divergence. 
For comparison with traditional metrics, we also include BLEU scores
measured on the WMT 2014 test set.

As we can see, the two PBMT systems fare very poorly on our
challenge set, especially in the morpho-syntactic and purely syntactic types. 
Their somewhat better handling of lexico-syntactic issues probably reflects the fact that
PBMT systems are naturally more attuned to lexical cues than to morphology or syntax.  
The two NMT systems are clear winners in all three categories. The GNMT
system is best overall
with a success rate of 68\%, likely due to the data
and architectural factors mentioned in section~\ref{ssec:neural}.\footnote{We cannot offer a full comparison 
with the pre-NMT Google system. However, in October 2016 we ran a smaller
35-sentence version of our challenge set on both the Google system and our
PBMT-1 system. The Google system only got 4 of those examples right (11.4\%) 
while our PBMT-1 got 6 right (17.1\%).}

WMT BLEU scores correlate poorly with challenge-set
performance. The large gap of 2.3 BLEU points between PBMT-1 and PBMT-2 corresponds to
only a 1\% gain on the challenge set,
while the small gap of 0.4 BLEU
between \mbox{PBMT-2} and NMT corresponds to a 21\% gain.

Inter-annotator agreement (final column in table~\ref{tab:results}) is
excellent overall, with all three annotators agreeing on almost 90\% of system
outputs. Syntactic divergences appear to be somewhat harder to judge than
other categories.

\begin{table*}[tb]
\centering
\begin{tabular}{llrrrr}
\hline
Category         & Subcategory                   & \# & PBMT-1   & NMT     & Google NMT    \\
\hline
Morpho-syntactic & Agreement across distractors  & 3  & 0\%   & 100\% & 100\% \\
                 & $\phantom{\ldots}$through control verbs         & 4  & 25\%  & 25\%  & 25\%  \\
                 & $\phantom{\ldots}$with coordinated target       & 3  & 0\%   & 100\% & 100\% \\
                 &$\phantom{\ldots}$with coordinated source       & 12 & 17\%  & 92\%  & 75\%  \\
                 & $\phantom{\ldots}$of past participles           & 4  & 25\%  & 75\%  & 75\%  \\
                 & Subjunctive mood              & 3  & 33\%  & 33\%  & 67\%  \\
\hline
Lexico-syntactic & Argument switch               & 3  & 0\%   & 0\%   & 0\%   \\
                 & Double-object verbs           & 3  & 33\%  & 67\%  & 100\% \\
                 & Fail-to                       & 3  & 67\%  & 100\% & 67\%  \\
                 & Manner-of-movement verbs      & 4  & 0\%   & 0\%   & 0\%   \\
                 & Overlapping subcat frames     & 5  & 60\%  & 100\% & 100\% \\
                 & NP-to-VP                      & 3  & 33\%  & 67\%  & 67\%  \\
                 & Factitives                    & 3  & 0\%   & 33\%  & 67\%  \\
                 & Noun compounds                & 9  & 67\%  & 67\%  & 78\%  \\
                 & Common idioms                 & 6  & 50\%  & 0\%   & 33\%  \\
                 & Syntactically flexible idioms & 2  & 0\%   & 0\%   & 0\%   \\
\hline
Syntactic        & Yes-no question syntax        & 3  & 33\%  & 100\% & 100\% \\
                 & Tag questions                 & 3  & 0\%   & 0\%   & 100\% \\
                 & Stranded preps                & 6  & 0\%   & 0\%   & 100\% \\
                 & Adv-triggered inversion       & 3  & 0\%   & 0\%   & 33\%  \\
                 & Middle voice                  & 3  & 0\%   & 0\%   & 0\%   \\
                 & Fronted should                & 3  & 67\%  & 33\%  & 33\%  \\
                 & Clitic pronouns               & 5  & 40\%  & 80\%  & 60\%  \\
                 & Ordinal placement             & 3  & 100\% & 100\% & 100\% \\
                 & Inalienable possession        & 6  & 50\%  & 17\%  & 83\%  \\
                 & Zero REL PRO                  & 3  & 0\%   & 33\%  & 100\%\\
\hline
\end{tabular}
\caption{
  Summary of scores by fine-grained categories. ``\#'' reports number of questions in each category,
  while the reported score is the percentage of questions for which the divergence was correctly bridged.
  For each question, the three human judgments were transformed into a single judgment
  by taking system outputs with two positive judgments as positive, and all others as negative.
}
\label{tab:fine}
\end{table*}

\subsection{Qualitative assessment of NMT}

We now turn to an analysis of the strengths and weaknesses of neural MT through the microscope of our divergence categorization system, hoping that this may help focus future research on key issues. In this discussion we ignore the results obtained by PBMT-2 and compare: a) the results obtained by PBMT-1 to those of NMT, both systems having been trained on the same dataset; and b) the results of these two systems with those of Google NMT which was trained on a much larger dataset. 

In the remainder of the present section we will refer to the sentences of our challenge set using the subcategory-based numbering scheme S1-S26
as assigned in Appendix~\ref{appendix:challenge_set}.
A summary of the category-wise performance of PBMT-1, NMT and Google NMT is provided in Table~\ref{tab:fine}.

\subsubsection*{Strengths of neural MT}

Overall, both neural MT systems do much better than PBMT-1 at bridging
divergences. In the case of  morpho-syntactic divergences, we observe a jump
from 16\% to 72\% in the case of our two local systems. This is mostly due to the NMT system's ability to deal with many of the more complex cases of subject-verb agrement: 

\begin{itemize}
\setlength\itemsep{-1pt}
\item {\em Distractors.} The subject's head noun agreement features get correctly passed 
to the verb phrase across intervening noun phrase complements (sentences S1a--c).
\item {\em Coordinated verb phrases.} Subject agreement marks are correctly distributed across the elements of such verb phrases (S3a--c).
\item {\em Coordinated subjects.} Much of the logic that is at stake in determining the agreement features of coordinated noun phrases (cf. our relevant description in section 3.1) appears to be correctly captured in the NMT translations of S4.
\item {\em Past participles.} Even though the rules governing French past participle agreement are notoriously difficult (especially after the ``avoir'' auxiliary), they are fairly well captured in the NMT translations of (S5b--e).
\end{itemize}
 
The NMT systems are also better at handling lexico-syntactic divergences. For example:
 
 \begin{itemize}
 \setlength\itemsep{-1pt}
\item {\em Double-object verbs.} There are no such verbs in French and the NMT systems perform the required adjustments flawlessly (sentences S8a--S8c).
\item {\em Overlapping subcat frames.} NMT systems manage to discriminate between an NP complement and a sentential complement starting with an NP: cf. {\em to know NP} versus {\em to know NP is VP} (S11b--e)
 \item {\em NP-to-VP complements.} These English infinitival complements often need to be rendered as finite clauses in French and the NMT systems are better at this task (S12a--c). 
\end{itemize}
 
Finally, NMT systems also turn out to better handle purely syntactic divergences. For example:

\begin{itemize}
 \setlength\itemsep{-1pt}
\item {\em Yes-no question syntax.} The differences between English and French yes-no question syntax are correctly bridged by the two NMT systems (S17a--c).
\item {\em French proclitics.} NMT systems are significantly better at transforming English pronouns into French proclitics, i.e. moving them before the main verb and case-inflecting them correctly (S23a--e). 
\item Finally, we note that the Google system manages to overcome several additional challenges. It correctly translates {\em tag questions} (S18a--c), constructions with {\em stranded prepositions} (S19a--f), most cases of the {\em inalienable possession} construction (S25a--e) as well as {\em zero relative pronouns} (S26a--c).
\end{itemize}

The large gap observed between the results of the in-house and Google NMT systems indicates that current neural MT systems are extremely data hungry. But given enough data, they can successfully tackle some challenges that are often thought of as extremely difficult. A case in point here is that of stranded prepositions (see discussion in section 3.3), in which we see the NMT model capture some instances of WH-movement, the textbook example of long-distance dependencies.

\subsubsection*{Weaknesses of neural MT}

In spite of its clear edge over PBMT, NMT is not without some serious
shortcomings. We already mentioned the degradation issue with long sentence which, by design, could not be observed with our challenge set. But an analysis of our results will reveal many other problems. Globally, we note that even using a
staggering quantity of data and a highly sophisticated NMT model, the Google
system fails to reach the 70\% mark on our challenge set. The fine-grained error categorization associated with the challenge set will help us single out precise areas where more research is needed. Here are some relevant observations.

{\em Incomplete generalizations}. In several cases where  partial results might suggest that NMT has correctly captured some basic generalization about linguistic data, further instances reveals that this is not fully the case.

\begin{itemize}
\item {\em Agreement logic.} The logic governing the agreement features of coordinated noun phrases (see section 3.1) has been mostly captured by the NMT systems (cf. the 12 sentences of S4), but there are some gaps. For example, the Google system runs into trouble with mixed-person subjects (sentences S4d1--3). 
\item {\em Subjunctive mood triggers.} While some subjunctive mood triggers are correctly registered (e.g. ``demander que'' and ``malheureux que''), the case of such a highly frequent subordinate conjunction as {\em provided that} $\rightarrow$ {\em à condition que} is somehow being missed (sentence S6a--c).
\item {\em Noun compounds.} The French translation of an English  compound {\em N$_{1}$ N$_{2}$} is usually of the form {\em N$_{2}$ Prep N$_{1}$}. For any given headnoun {\em N$_{2}$} the correct preposition {\em Prep} depends on the semantic class of {\em N$_{1}$}. For example {\em steel/ceramic/plastic knife $\rightarrow$ couteau \textbf{en} acier/céramique/plastique} but {\em butter/meat/steak knife $\rightarrow$ couteau \textbf{à} beurre/viande/steak}. Given that neural models are known to perform some semantic generalizations, we find their performance disappointing on our compound noun examples (S14a--i).
\item The so-called French ``inalienable possession'' construction arises when an agent performs an action on one of her body parts, e.g. {\em I brushed my teeth}. The French translation will normally replace the possessive article with a definite one and introduce a reflexive pronoun, e.g. {\em Je me suis brossé les dents} ('I brushed myself the teeth'). In our dataset, the Google system gets this right for examples in the first and third persons (sentences S25a,b) but fails to do the same with the example in the second person (sentence S25c).
\end{itemize}

Then there are also phenomena that current NMT systems, even with massive amounts of data, appear to be completely missing:
\begin{itemize}
\setlength\itemsep{-1pt}
\item {\em Common and syntactically flexible idioms}. While PBMT-1 produces an acceptable translation for half of the idiomatic expressions of S15 and S16, the local NMT system misses them all and the Google system does barely better. NMT systems appear to be short on raw memorization capabilities.
\item {\em Control verbs}. Two different classes of verbs can govern a subject NP, an object NP plus an infinitival complement. With verbs of the ``object-control'' class (e.g. ``persuade''), the object of the verb is understood as the semantic subject of the infinitive. But with those of the ``subject-control'' class (e.g. ``promise''), it is rather the subject of the verb which plays that semantic role. None of the systems tested here appear to get a grip on subject control cases, as evidenced by the lack of correct feminine agreement on the French adjectives in sentences S2b--d.
\item {\em Argument switching verbs}. All systems tested here mistranslate sentences S7a--c by failing to perform the required argument switch: {\em NP$_1$ misses NP$_2$} $\rightarrow$ {\em NP$_2$ manque à NP$_1$}.
\item {\em Crossing movement verbs}. None of the systems managed to correctly restructure the regular manner-of-movement verbs e.g. {\em swim across X} $\rightarrow$ {\em traverser X à la nage} in sentences S10a-c. Unsurprisingly, all systems also fail on the even harder example S10d, in which the ``nonce verb'' {\em guitared} is a spontaneous derivation from the noun {\em guitar} being cast as an ad hoc manner-of-movement verb. \footnote{ On the concept of nonce word, see https://en.wikipedia.org/wiki/Nonce\_word.}
\item {\em Middle voice}. None of the systems tested here were able to recast the English ``generic passive'' of S21a--c into the expected French ``middle voice'' pronominal construction.

\end{itemize}

\section{Conclusions}

We have presented a radically different kind of evaluation for MT
systems: the use of challenge sets designed to stress-test MT systems on ``hard'' linguistic material, while providing a fine-grained linguistic classification of their successes and failures. This approach is not meant to replace our community's traditional evaluation tools but to supplement them.


Our proposed error categorization scheme  makes it possible to bring to light different strengths and weaknesses of PBMT and neural MT. With the exception of idiom processing, in all cases where a clear difference 
was observed it turned out to be in favor of neural MT. A key factor in NMT's superiority appears to be its ability to overcome many limitations of $n$-gram language modeling. This is clearly at play in dealing with subject-verb agreement, double-object verbs, overlapping subcategorization frames and last but not least, the pinnacle of Chomskyan linguistics, WH-movement (in this case, stranded prepositions).


But our challenge set also brings to light some important shortcomings of current neural MT, regardless of the massive amounts of training data it may have been fed. As may have been already known or suspected, NMT systems struggle with the translation of idiomatic phrases.
Perhaps more interestingly, we notice that neural MT's impressive generalizations still seem somewhat brittle. 
For example, the NMT system can appear to have mastered the rules governing subject-verb agreement or inalienable possession in French, only to trip over a rather obvious instantiation of those rules. 
Probing where these boundaries are, and how they relate to the neural system's training data and architecture is an obvious next step.


\section{Future Work}
It is our hope that the insights derived from our challenge set evaluation will help inspire future MT research,
and call attention to the fact that even ``easy'' language pairs like English--French still have many linguistic issues left to be resolved.
But there are also several ways to improve and expand upon our challenge set approach itself.

First, though our human judgments of output sentences allowed us to precisely
assess the phenomena of interest, this approach is not scalable to large sets,
and requires access to native speakers in order to replicate the evaluation.  It
would be interesting to see whether similar scores could be achieved through
automatic means.  The existence of human judgments for this set provides a
gold-standard by which proposed automatic judgments may be meta-evaluated.

Second, the construction of such a challenge set requires in-depth knowledge 
of the structural divergences between the two languages of interest.
A method to automatically create such a challenge set for a new language pair would be extremely useful.
One could imagine approaches that search for divergences, indicated by atypical output configurations, 
or perhaps by a system's inability to reproduce a reference from its own training data.
Localizing a divergence within a difficult sentence pair would be another useful subtask.

Finally, we would like to explore how to train an MT system 
to improve its performance on these divergence phenomena.
This could take the form of designing a curriculum to demonstrate a particular divergence to the machine,
or altering the network structure to capture such generalizations.

\section*{Acknowledgments}

We would like to thank Cyril Goutte, Eric Joanis and Michel Simard,  who
graciously spent the time required to rate the output of four different MT
systems on our challenge sentences. We also thank Roland Kuhn for valuable
discussions, and comments on an earlier version of the paper.

\bibliography{acl2017}
\bibliographystyle{emnlp_natbib}

\appendix
\clearpage


\section{Instructions to Annotators}
\label{appendix:annotator_instructions}
The following instructions were provided to annotators:

{\em
You will be presented with 108 short English sentences and the French translations produced for them by each of four different machine translation systems. You will not be asked to provide an overall rating for the machine-translated sentences. Rather, you will be asked to determine whether or not a highly specific aspect of the English sentence is correctly rendered in each of the different translations. Each English sentence will be accompanied with a yes-no question which precisely specifies the targeted element for the associated translations. For example, you may be asked to determine whether or not the main verb phrase of the translation is in correct grammatical agreement with its subject.

In order to facilitate this process, each English sentence will also be provided with a French reference (human) translation in which the particular elements that support a yes answer (in our example, the correctly agreeing verb phrase) will be highlighted. Your answer should be “yes” if the question can be answered positively and “no” otherwise. Note that this means that any translation error which is unrelated to the question at hand should be disregarded. Using the same example: as long as the verb phrase agrees correctly with its subject, it does not matter whether or not the verb is correctly chosen, is in the right tense, etc. And of course, it does not matter if unrelated parts of the translation are wrong.

 In most cases you should be able to quickly determine a positive or negative
 answer. However, there may be cases in which the system has come up with a
 translation that just does not contain the phenomenon targeted by the
 associated question. In such cases, and only in such cases, you should choose
 “not applicable” regardless of whether or not the translation is correct.
}

\section{Challenge Set}
\label{appendix:challenge_set}

We include a rendering of our challenge set in the pages that follow, along
with system output for the PBMT-1, NMT and Google systems.\footnote{A machine-readable version is provided in the file {\tt Challenge\_set-v2hA.json} in the supplemental materials.}
Sentences are
grouped by linguistic category and subcategory.  For convenience, we also include a reference translation, which is a manually-crafted translation that is designed to be the most straightforward solution to the divergence problem at hand.
Needless to say, this reference translation is seldom the only acceptable solution to the targeted divergence problem.
Our judges were provided these references, but were instructed to use their knowledge of French to judge whether the divergence was correctly bridged,
regardless of the translation's similarity to the reference.


In all translations, the locus of the targeted divergence is highlighted in boldface and it is specifically on that portion that our annotators were asked to provide a judgment.
For each system output, we provide a summary of our annotator's judgments on its handling of the phenomenon of interest.
We label the translation with a \cmark\ if two or more annotators judged the divergence to be correctly bridged, and with an \xmark\ otherwise.
 
We also release a machine-readable version of this same data,
including all of the individual judgments,
in the hope that others will find interesting new uses for it.

\include{Challenge_set-v2hA}


\end{document}

%% file: Challenge_set-v2hA.tex
\begin{table*}
\begin{tabular}{cll}
\\\multicolumn{3}{l}{\large \textbf{Morpho-Syntactic}}\\
\\\multicolumn{3}{l}{\textbf{S-V agreement, across distractors}}\\
\hline
\multicolumn{3}{p{15cm}}{Is subject-verb agrement correct? (Possible interference from distractors between the subject's head and the verb).}\\
\hline
S1a & Source & The repeated calls from his mother \textbf{should} have alerted us. \\
& Ref & Les appels répétés de sa mère \textbf{auraient} dû nous alerter. \\
& PBMT-1 & Les appels répétés de sa mère aurait dû nous a alertés. \xmark\\
& NMT & Les appels répétés de sa mère devraient nous avoir alertés. \cmark\\
& Google & Les appels répétés de sa mère auraient dû nous alerter. \cmark\\
\hline
S1b & Source & The sudden noise in the upper rooms \textbf{should} have alerted us. \\
& Ref & Le bruit soudain dans les chambres supérieures \textbf{aurait} dû nous alerter. \\
& PBMT-1 & Le bruit soudain dans les chambres supérieures auraient dû nous a alertés. \xmark\\
& NMT & Le bruit soudain dans les chambres supérieures devrait nous avoir alerté. \cmark\\
& Google & Le bruit soudain dans les chambres supérieures devrait nous avoir alerté. \cmark\\
\hline
S1c & Source & Their repeated failures to report the problem \textbf{should} have alerted us. \\
& Ref & Leurs échecs répétés à signaler le problème \textbf{auraient} dû nous alerter. \\
& PBMT-1 & Leurs échecs répétés de signaler le problème aurait dû nous a alertés. \xmark\\
& NMT & Leurs échecs répétés pour signaler le problème devraient nous avoir alertés. \cmark\\
& Google & Leur échec répété à signaler le problème aurait dû nous alerter. \cmark\\
\\\multicolumn{3}{l}{\textbf{S-V agreement, through control verbs}}\\
\hline
\multicolumn{3}{p{15cm}}{Does the flagged adjective agree correctly with its subject? (Subject-control versus object-control verbs).}\\
\hline
S2a & Source & She asked her brother not to be \textbf{arrogant}. \\
& Ref & Elle a demandé à son frère de ne pas se montrer \textbf{arrogant}. \\
& PBMT-1 & Elle a demandé à son frère de ne pas être arrogant. \cmark\\
& NMT & Elle a demandé à son frère de ne pas être arrogant. \cmark\\
& Google & Elle a demandé à son frère de ne pas être arrogant. \cmark\\
\hline
S2b & Source & She promised her brother not to be \textbf{arrogant}. \\
& Ref & Elle a promis à son frère de ne pas être \textbf{arrogante}. \\
& PBMT-1 & Elle a promis son frère à ne pas être arrogant. \xmark\\
& NMT & Elle a promis à son frère de ne pas être arrogant. \xmark\\
& Google & Elle a promis à son frère de ne pas être arrogant. \xmark\\
\hline
S2c & Source & She promised her doctor to remain \textbf{active} after retiring. \\
& Ref & Elle a promis à son médecin de demeurer \textbf{active} après s'être retirée. \\
& PBMT-1 & Elle a promis son médecin pour demeurer actif après sa retraite. \xmark\\
& NMT & Elle a promis à son médecin de rester actif après sa retraite. \xmark\\
& Google & Elle a promis à son médecin de rester actif après sa retraite. \xmark\\
\hline
S2d & Source & My mother promised my father to be more \textbf{prudent} on the road. \\
& Ref & Ma mère a promis à mon père d'être plus \textbf{prudente} sur la route. \\
& PBMT-1 & Ma mère, mon père a promis d'être plus prudent sur la route. \xmark\\
& NMT & Ma mère a promis à mon père d'être plus prudent sur la route. \xmark\\
& Google & Ma mère a promis à mon père d'être plus prudent sur la route. \xmark\\
\end{tabular}
\end{table*}
\begin{table*}
\begin{tabular}{cll}
\\\multicolumn{3}{l}{\textbf{S-V agreement, coordinated targets}}\\
\hline
\multicolumn{3}{p{15cm}}{Do the marked verbs/adjective agree correctly with their subject? (Agreement distribution over coordinated predicates)}\\
\hline
S3a & Source & The woman was very \textbf{tall} and extremely \textbf{strong}. \\
& Ref & La femme était très \textbf{grande} et extrêmement \textbf{forte}. \\
& PBMT-1 & La femme était très gentil et extrêmement forte. \xmark\\
& NMT & La femme était très haute et extrêmement forte. \cmark\\
& Google & La femme était très grande et extrêmement forte. \cmark\\
\hline
S3b & Source & Their politicians were more \textbf{ignorant} than \textbf{stupid}. \\
& Ref & Leurs politiciens étaient plus \textbf{ignorants} que \textbf{stupides}. \\
& PBMT-1 & Les politiciens étaient plus ignorants que stupide. \xmark\\
& NMT & Leurs politiciens étaient plus ignorants que stupides. \cmark\\
& Google & Leurs politiciens étaient plus ignorants que stupides. \cmark\\
\hline
S3c & Source & We \textbf{shouted} an insult and \textbf{left} abruptly. \\
& Ref & Nous \textbf{avons} lancé une insulte et nous \textbf{sommes} partis brusquement. \\
& PBMT-1 & Nous avons crié une insulte et a quitté abruptement. \xmark\\
& NMT & Nous avons crié une insulte et nous avons laissé brusquement. \cmark\\
& Google & Nous avons crié une insulte et nous sommes partis brusquement. \cmark\\
\\\multicolumn{3}{l}{\textbf{S-V agreement, feature calculus on coordinated source}}\\
\hline
\multicolumn{3}{p{15cm}}{Do the marked verbs/adjective agree correctly with their subject? (Masculine singular ET masculine singular yields masculine plural).}\\
\hline
S4a1 & Source & The cat and the dog \textbf{should} be \textbf{watched}. \\
& Ref & Le chat et le chien \textbf{devraient} être \textbf{surveillés}. \\
& PBMT-1 & Le chat et le chien doit être regardée. \xmark\\
& NMT & Le chat et le chien doivent être regardés. \cmark\\
& Google & Le chat et le chien doivent être surveillés. \cmark\\
\hline
S4a2  & Source & My father and my brother \textbf{will} be \textbf{happy} tomorrow. \\
& Ref & Mon père et mon frère \textbf{seront} \textbf{heureux} demain. \\
& PBMT-1 & Mon père et mon frère sera heureux de demain. \xmark\\
& NMT & Mon père et mon frère seront heureux demain. \cmark\\
& Google & Mon père et mon frère seront heureux demain. \cmark\\
\hline
S4a3 & Source & My book and my pencil \textbf{could} be \textbf{stolen}. \\
& Ref & Mon livre et mon crayon \textbf{pourraient} être \textbf{volés}. \\
& PBMT-1 & Mon livre et mon crayon pourrait être volé. \xmark\\
& NMT & Mon livre et mon crayon pourraient être volés. \cmark\\
& Google & Mon livre et mon crayon pourraient être volés. \cmark\\
\hline
\multicolumn{3}{p{15cm}}{Do the marked verbs/adjectives agree correctly with their subject? (Feminine singular ET feminine singular yields feminine plural).}\\
\hline
S4b1 & Source & The cow and the hen \textbf{must} be \textbf{fed}. \\
& Ref & La vache et la poule \textbf{doivent} être \textbf{nourries}. \\
& PBMT-1 & La vache et de la poule doivent être nourris. \xmark\\
& NMT & La vache et la poule doivent être alimentées. \cmark\\
& Google & La vache et la poule doivent être nourries. \cmark\\
\end{tabular}
\end{table*}
\begin{table*}
\begin{tabular}{cll}
\hline
S4b2 & Source & My mother and my sister \textbf{will be happy} tomorrow. \\
& Ref & Ma mère et ma sœur \textbf{seront heureuses} demain. \\
& PBMT-1 & Ma mère et ma sœur sera heureux de demain. \xmark\\
& NMT & Ma mère et ma sœur seront heureuses demain. \cmark\\
& Google & Ma mère et ma sœur seront heureuses demain. \cmark\\
\hline
S4b3 & Source & My shoes and my socks \textbf{will} be \textbf{found}. \\
& Ref & Mes chaussures et mes chaussettes \textbf{seront} \textbf{retrouvées}. \\
& PBMT-1 & Mes chaussures et mes chaussettes sera trouvé. \xmark\\
& NMT & Mes chaussures et mes chaussettes seront trouvées. \cmark\\
& Google & Mes chaussures et mes chaussettes seront trouvées. \cmark\\
\hline
\multicolumn{3}{p{15cm}}{Do the marked verbs/adjectives agree correctly with their subject? (Masculine singular ET feminine singular yields masculine plural.)}\\
\hline
S4c1 & Source & The dog and the cow \textbf{are} \textbf{nervous}. \\
& Ref & Le chien et la vache \textbf{sont} \textbf{nerveux}. \\
& PBMT-1 & Le chien et la vache sont nerveux. \cmark\\
& NMT & Le chien et la vache sont nerveux. \cmark\\
& Google & Le chien et la vache sont nerveux. \cmark\\
\hline
S4c2 & Source & My father and my mother will be happy tomorrow. \\
& Ref & Mon père et ma mère \textbf{seront} \textbf{heureux} demain. \\
& PBMT-1 & Mon père et ma mère se fera un plaisir de demain. \xmark\\
& NMT & Mon père et ma mère seront heureux demain. \cmark\\
& Google & Mon père et ma mère seront heureux demain. \cmark\\
\hline
S4c3 & Source & My refrigerator and my kitchen table \textbf{were} \textbf{stolen}. \\
& Ref & Mon réfrigérateur et ma table de cuisine \textbf{ont} été \textbf{volés}. \\
& PBMT-1 & Mon réfrigérateur et ma table de cuisine ont été volés. \cmark\\
& NMT & Mon réfrigérateur et ma table de cuisine ont été volés. \cmark\\
& Google & Mon réfrigérateur et ma table de cuisine ont été volés. \cmark\\
\hline
\multicolumn{3}{p{15cm}}{Do the marked verbs/adjectives agree correctly with their subject? (Smallest coordinated grammatical person wins.)}\\
\hline
S4d1 & Source & Paul and I \textbf{could} easily be \textbf{convinced} to join you. \\
& Ref & Paul et moi \textbf{pourrions} facilement être \textbf{convaincus} de se joindre à vous. \\
& PBMT-1 & Paul et je pourrais facilement être persuadée de se joindre à vous. \xmark\\
& NMT & Paul et moi avons facilement pu être convaincus de vous rejoindre. \cmark\\
& Google & Paul et moi pourrait facilement être convaincu de vous rejoindre. \xmark\\
\hline
S4d2 & Source & You and he \textbf{could} be \textbf{surprised} by her findings. \\
& Ref & Vous et lui \textbf{pourriez} être \textbf{surpris} par ses découvertes. \\
& PBMT-1 & Vous et qu'il pouvait être surpris par ses conclusions. \xmark\\
& NMT & Vous et lui pourriez être surpris par ses conclusions. \cmark\\
& Google & Vous et lui pourrait être surpris par ses découvertes. \xmark\\
\end{tabular}
\end{table*}
\begin{table*}
\begin{tabular}{cll}
\hline
S4d3 & Source & We and they \textbf{are} on different courses. \\
& Ref & Nous et eux \textbf{sommes} sur des trajectoires différentes. \\
& PBMT-1 & Nous et ils sont en cours de différents. \xmark\\
& NMT & Nous et nous sommes sur des parcours différents. \xmark\\
& Google & Nous et ils sont sur des parcours différents. \xmark\\
\\\multicolumn{3}{l}{\textbf{S-V agreement, past participles}}\\
\hline
\multicolumn{3}{p{15cm}}{Are the agreement marks of the flagged participles the correct ones? (Past participle placed after auxiliary AVOIR agrees with verb object iff object precedes auxiliary. Otherwise participle is in masculine singular form).}\\
\hline
S5a & Source & The woman who \textbf{saw} a mouse in the corridor is charming. \\
& Ref & La femme qui a \textbf{vu} une souris dans le couloir est charmante. \\
& PBMT-1 & La femme qui a vu une souris dans le couloir est charmante. \cmark\\
& NMT & La femme qui a vu une souris dans le couloir est charmante. \cmark\\
& Google & La femme qui a vu une souris dans le couloir est charmante. \cmark\\
\hline
S5b & Source & The woman that your brother \textbf{saw} in the corridor is charming. \\
& Ref & La femme que votre frère a \textbf{vue} dans le couloir est charmante. \\
& PBMT-1 & La femme que ton frère a vu dans le couloir est charmante. \xmark\\
& NMT & La femme que votre frère a vu dans le corridor est charmante. \xmark\\
& Google & La femme que votre frère a vue dans le couloir est charmante. \cmark\\
\hline
S5c & Source & The house that John has \textbf{visited} is crumbling. \\
& Ref & La maison que John a \textbf{visitée} tombe en ruines. \\
& PBMT-1 & La maison que John a visité est en train de s'écrouler. \xmark\\
& NMT & La maison que John a visitée est en train de s'effondrer. \cmark\\
& Google & La maison que John a visité est en ruine. \xmark\\
\hline
S5d & Source & John sold the car that he had \textbf{won} in a lottery. \\
& Ref & John a vendu la voiture qu'il avait \textbf{gagnée} dans une loterie. \\
& PBMT-1 & John a vendu la voiture qu'il avait gagné à la loterie. \xmark\\
& NMT & John a vendu la voiture qu'il avait gagnée dans une loterie. \cmark\\
& Google & John a vendu la voiture qu'il avait gagnée dans une loterie. \cmark\\
\\\multicolumn{3}{l}{\textbf{Subjunctive mood}}\\
\hline
\multicolumn{3}{p{15cm}}{Is the flagged verb in the correct mood? (Certain triggering verbs, adjectives or subordinate conjunctions, induce the subjunctive mood in the subordinate clause that they govern). }\\
\hline
S6a & Source & He will come provided that you \textbf{come} too. \\
& Ref & Il viendra à condition que vous \textbf{veniez} aussi. \\
& PBMT-1 & Il viendra à condition que vous venez aussi. \xmark\\
& NMT & Il viendra lui aussi que vous le faites. \xmark\\
& Google & Il viendra à condition que vous venez aussi. \xmark\\
\hline
S6b & Source & It is unfortunate that he is not \textbf{coming} either. \\
& Ref & Il est malheureux qu'il ne \textbf{vienne} pas non plus. \\
& PBMT-1 & Il est regrettable qu'il n'est pas non plus à venir. \xmark\\
& NMT & Il est regrettable qu'il ne soit pas non plus. \xmark\\
& Google & Il est malheureux qu'il ne vienne pas non plus. \cmark\\
\end{tabular}
\end{table*}
\begin{table*}
\begin{tabular}{cll}
\hline
S6c & Source & I requested that families not \textbf{be} separated. \\
& Ref & J'ai demandé que les familles ne \textbf{soient} pas séparées. \\
& PBMT-1 & J'ai demandé que les familles ne soient pas séparées. \cmark\\
& NMT & J'ai demandé que les familles ne soient pas séparées. \cmark\\
& Google & J'ai demandé que les familles ne soient pas séparées. \cmark\\
\\\multicolumn{3}{l}{\large \textbf{Lexico-Syntactic}}\\
\\\multicolumn{3}{l}{\textbf{Argument switch}}\\
\hline
\multicolumn{3}{p{15cm}}{Are the experiencer and the object of the ``missing'' situation correctly preserved in the French translation? (Argument switch).}\\
\hline
S7a & Source & \textbf{Mary} sorely misses \textbf{Jim}. \\
& Ref & \textbf{Jim} manque cruellement \textbf{à Mary}. \\
& PBMT-1 & Marie manque cruellement de Jim. \xmark\\
& NMT & Mary a lamentablement manqué de Jim. \xmark\\
& Google & Mary manque cruellement à Jim. \xmark\\
\hline
S7b & Source & \textbf{My sister} is really missing \textbf{New York.} \\
& Ref & \textbf{New York} manque beaucoup \textbf{à ma sœur}. \\
& PBMT-1 & Ma sœur est vraiment absent de New York. \xmark\\
& NMT & Ma sœur est vraiment manquante à New York. \xmark\\
& Google & Ma sœur manque vraiment New York. \xmark\\
\hline
S7c & Source & What \textbf{he} misses most is \textbf{his dog}. \\
& Ref & Ce qui \textbf{lui} manque le plus, c'est \textbf{son chien}. \\
& PBMT-1 & Ce qu'il manque le plus, c'est son chien. \xmark\\
& NMT & Ce qu'il manque le plus, c'est son chien. \xmark\\
& Google & Ce qu'il manque le plus, c'est son chien. \xmark\\
\\\multicolumn{3}{l}{\textbf{Double-object verbs}}\\
\hline
\multicolumn{3}{p{15cm}}{Are ``gift'' and ``recipient'' arguments correctly rendered in French? (English double-object constructions)}\\
\hline
S8a & Source & John gave \textbf{his wonderful wife} a nice present. \\
& Ref & John a donné un beau présent \textbf{à sa merveilleuse épouse}. \\
& PBMT-1 & John a donné sa merveilleuse femme un beau cadeau. \xmark\\
& NMT & John a donné à sa merveilleuse femme un beau cadeau. \cmark\\
& Google & John a donné à son épouse merveilleuse un présent gentil. \cmark\\
\hline
S8b & Source & John told \textbf{the kids} a nice story. \\
& Ref & John a raconté une belle histoire \textbf{aux enfants}. \\
& PBMT-1 & John a dit aux enfants une belle histoire. \cmark\\
& NMT & John a dit aux enfants une belle histoire. \cmark\\
& Google & John a raconté aux enfants une belle histoire. \cmark\\
\hline
S8c & Source & John sent \textbf{his mother} a nice postcard. \\
& Ref & John a envoyé une belle carte postale \textbf{à sa mère}. \\
& PBMT-1 & John a envoyé sa mère une carte postale de nice. \xmark\\
& NMT & John a envoyé sa mère une carte postale de nice. \xmark\\
& Google & John envoya à sa mère une belle carte postale. \cmark\\
\end{tabular}
\end{table*}
\begin{table*}
\begin{tabular}{cll}
\\\multicolumn{3}{l}{\textbf{Fail to}}\\
\hline
\multicolumn{3}{p{15cm}}{Is the meaning of ``fail to'' correctly rendered in the French translation? }\\
\hline
S9a & Source & John \textbf{failed to} see the relevance of this point. \\
& Ref & John \textbf{n'a pas} vu la pertinence de ce point. \\
& PBMT-1 & John a omis de voir la pertinence de ce point. \xmark\\
& NMT & John n'a pas vu la pertinence de ce point. \cmark\\
& Google & John a omis de voir la pertinence de ce point. \xmark\\
\hline
S9b & Source & He failed to respond. \\
& Ref & Il \textbf{n'a pas répondu}. \\
& PBMT-1 & Il n'a pas réussi à répondre. \cmark\\
& NMT & Il n'a pas répondu. \cmark\\
& Google & Il n'a pas répondu. \cmark\\
\hline
S9c & Source & Those who fail to comply with this requirement will be penalized. \\
& Ref & Ceux qui \textbf{ne} se conforment \textbf{pas} à cette exigence seront pénalisés. \\
& PBMT-1 & Ceux qui ne se conforment pas à cette obligation seront pénalisés. \cmark\\
& NMT & Ceux qui ne se conforment pas à cette obligation seront pénalisés. \cmark\\
& Google & Ceux qui ne respectent pas cette exigence seront pénalisés. \cmark\\
\\\multicolumn{3}{l}{\textbf{Manner-of-movement verbs}}\\
\hline
\multicolumn{3}{p{15cm}}{Is the movement action expressed in the English source correctly rendered in French? (Manner-of-movement verbs with path argument may need to be rephrased in French).}\\
\hline
S10a & Source & John would like to \textbf{swim across} the river. \\
& Ref & John aimerait \textbf{traverser} la rivière \textbf{à la nage}. \\
& PBMT-1 & John aimerait nager dans la rivière. \xmark\\
& NMT & John aimerait nager à travers la rivière. \xmark\\
& Google & John aimerait nager à travers la rivière. \xmark\\
\hline
S10b & Source & They \textbf{ran into} the room. \\
& Ref & Ils \textbf{sont entrés} dans la chambre \textbf{à la course}. \\
& PBMT-1 & Ils ont couru dans la chambre. \xmark\\
& NMT & Ils ont couru dans la pièce. \xmark\\
& Google & Ils coururent dans la pièce. \xmark\\
\hline
S10c & Source & The man \textbf{ran out of} the park. \\
& Ref & L'homme \textbf{est sorti du} parc \textbf{en courant}. \\
& PBMT-1 & L'homme a manqué du parc. \xmark\\
& NMT & L'homme s'enfuit du parc. \xmark\\
& Google & L'homme sortit du parc. \xmark\\
\hline
\multicolumn{3}{p{15cm}}{Hard example featuring spontaneous noun-to-verb derivation (``nonce verb'').}\\
\hline
S10d & Source & John \textbf{guitared his way} to San Francisco. \\
& Ref & John \textbf{s'est rendu} jusqu'à San Francisco \textbf{en jouant de la guitare}. \\
& PBMT-1 & John guitared son chemin à San Francisco. \xmark\\
& NMT & John guitared sa route à San Francisco. \xmark\\
& Google & John a guité son chemin à San Francisco. \xmark\\
\end{tabular}
\end{table*}
\begin{table*}
\begin{tabular}{cll}
\\\multicolumn{3}{l}{\textbf{Overlapping subcat frames}}\\
\hline
\multicolumn{3}{p{15cm}}{Is the French verb for ``know'' correctly chosen? (Choice between ``savoir''/``connaître'' depends on syntactic nature of its object)}\\
\hline
S11a & Source & Paul \textbf{knows} that this is a fact. \\
& Ref & Paul \textbf{sait que} c'est un fait. \\
& PBMT-1 & Paul sait que c'est un fait. \cmark\\
& NMT & Paul sait que c'est un fait. \cmark\\
& Google & Paul sait que c'est un fait. \cmark\\
\hline
S11b & Source & Paul \textbf{knows} this story. \\
& Ref & Paul \textbf{connaît} cette histoire. \\
& PBMT-1 & Paul connaît cette histoire. \cmark\\
& NMT & Paul connaît cette histoire. \cmark\\
& Google & Paul connaît cette histoire. \cmark\\
\hline
S11c & Source & Paul \textbf{knows} this story is hard to believe. \\
& Ref & Paul \textbf{sait que} cette histoire est difficile à croire. \\
& PBMT-1 & Paul connaît cette histoire est difficile à croire. \xmark\\
& NMT & Paul sait que cette histoire est difficile à croire. \cmark\\
& Google & Paul sait que cette histoire est difficile à croire. \cmark\\
\hline
S11d & Source & He \textbf{knows} my sister will not take it. \\
& Ref & Il \textbf{sait que} ma soeur ne le prendra pas. \\
& PBMT-1 & Il sait que ma soeur ne prendra pas. \cmark\\
& NMT & Il sait que ma soeur ne le prendra pas. \cmark\\
& Google & Il sait que ma soeur ne le prendra pas. \cmark\\
\hline
S11e & Source & My sister \textbf{knows} your son is reliable. \\
& Ref & Ma sœur \textbf{sait que} votre fils est fiable. \\
& PBMT-1 & Ma soeur connaît votre fils est fiable. \xmark\\
& NMT & Ma sœur sait que votre fils est fiable. \cmark\\
& Google & Ma sœur sait que votre fils est fiable. \cmark\\
\\\multicolumn{3}{l}{\textbf{NP to VP}}\\
\hline
\multicolumn{3}{p{15cm}}{Is the English ``NP to VP'' complement correctly rendred in the French translation? (Sometimes one needs to translate this structure as a finite clause).}\\
\hline
S12a & Source & John believes \textbf{Bill to be dishonest}. \\
& Ref & John croit \textbf{que Bill est malhonnête}. \\
& PBMT-1 & John estime que le projet de loi soit malhonnête. \cmark\\
& NMT & John croit que le projet de loi est malhonnête. \cmark\\
& Google & John croit que Bill est malhonnête. \cmark\\
\hline
S12b & Source & He liked \textbf{his father to tell him stories}. \\
& Ref & Il aimait \textbf{que son père lui raconte des histoires}. \\
& PBMT-1 & Il aimait son père pour lui raconter des histoires. \xmark\\
& NMT & Il aimait son père pour lui raconter des histoires. \xmark\\
& Google & Il aimait son père à lui raconter des histoires. \xmark\\
\end{tabular}
\end{table*}
\begin{table*}
\begin{tabular}{cll}
\hline
S12c & Source & She wanted \textbf{her mother to let her go}. \\
& Ref & Elle voulait \textbf{que sa mère la laisse partir}. \\
& PBMT-1 & Elle voulait que sa mère de lui laisser aller. \xmark\\
& NMT & Elle voulait que sa mère la laisse faire. \cmark\\
& Google & Elle voulait que sa mère la laisse partir. \cmark\\
\\\multicolumn{3}{l}{\textbf{Factitives}}\\
\hline
\multicolumn{3}{p{15cm}}{Is the English verb correctly rendered in the French translation? (Agentive use of some French verbs require embedding under ``faire'').}\\
\hline
S13a & Source & John \textbf{cooked} a big chicken. \\
& Ref & John a \textbf{fait cuire} un gros poulet. \\
& PBMT-1 & John cuit un gros poulet. \xmark\\
& NMT & John cuit un gros poulet. \xmark\\
& Google & John a fait cuire un gros poulet. \cmark\\
\hline
S13b & Source & John \textbf{melted} a lot of ice. \\
& Ref & John a \textbf{fait fondre} beaucoup de glace. \\
& PBMT-1 & John fondu a lot of ice. \xmark\\
& NMT & John a fondu beaucoup de glace. \xmark\\
& Google & John a fondu beaucoup de glace. \xmark\\
\hline
S13c & Source & She likes to \textbf{grow} flowers. \\
& Ref & Elle aime \textbf{faire pousser} des fleurs. \\
& PBMT-1 & Elle aime à se développer des fleurs. \xmark\\
& NMT & Elle aime à cultiver des fleurs. \cmark\\
& Google & Elle aime faire pousser des fleurs. \cmark\\
\\\multicolumn{3}{l}{\textbf{Noun Compounds}}\\
\hline
\multicolumn{3}{p{15cm}}{Is the English nominal compound rendered with the right preposition in the French translation?}\\
\hline
S14a & Source & Use the meat knife. \\
& Ref & Utilisez le couteau \textbf{à} viande. \\
& PBMT-1 & Utilisez le couteau de viande. \xmark\\
& NMT & Utilisez le couteau à viande. \cmark\\
& Google & Utilisez le couteau à viande. \cmark\\
\hline
S14b & Source & Use the butter knife. \\
& Ref & Utilisez le couteau \textbf{à} beurre. \\
& PBMT-1 & Utilisez le couteau à beurre. \cmark\\
& NMT & Utilisez le couteau au beurre. \xmark\\
& Google & Utilisez le couteau à beurre. \cmark\\
\hline
S14c & Source & Use the steak knife. \\
& Ref & Utilisez le couteau \textbf{à} steak. \\
& PBMT-1 & Utilisez le steak couteau. \xmark\\
& NMT & Utilisez le couteau à steak. \cmark\\
& Google & Utilisez le couteau de steak. \xmark\\
\end{tabular}
\end{table*}
\begin{table*}
\begin{tabular}{cll}
\hline
S14d & Source & Clean the water filter. \\
& Ref & Nettoyez le filtre \textbf{à} eau. \\
& PBMT-1 & Nettoyez le filtre à eau. \cmark\\
& NMT & Nettoyez le filtre à eau. \cmark\\
& Google & Nettoyez le filtre à eau. \cmark\\
\hline
S14e & Source & Clean the juice filter. \\
& Ref & Nettoyez le filtre \textbf{à} jus. \\
& PBMT-1 & Nettoyez le filtre de jus. \xmark\\
& NMT & Nettoyez le filtre de jus. \xmark\\
& Google & Nettoyez le filtre à jus. \cmark\\
\hline
S14f & Source & Clean the tea filter. \\
& Ref & Nettoyez le filtre \textbf{à} thé. \\
& PBMT-1 & Nettoyez le filtre à thé. \cmark\\
& NMT & Nettoyez le filtre de thé. \xmark\\
& Google & Nettoyez le filtre à thé. \cmark\\
\hline
S14g & Source & Clean the cloth filter. \\
& Ref & Nettoyez le filtre \textbf{en} tissu. \\
& PBMT-1 & Nettoyez le filtre en tissu. \cmark\\
& NMT & Nettoyez le filtre en tissu. \cmark\\
& Google & Nettoyez le filtre en tissu. \cmark\\
\hline
S14h & Source & Clean the metal filter. \\
& Ref & Nettoyez le filtre \textbf{en} métal. \\
& PBMT-1 & Nettoyez le filtre en métal. \cmark\\
& NMT & Nettoyez le filtre en métal. \cmark\\
& Google & Nettoyez le filtre métallique. \cmark\\
\hline
S14i & Source & Clean the paper filter. \\
& Ref & Nettoyez le filtre \textbf{en} papier. \\
& PBMT-1 & Nettoyez le filtre en papier. \cmark\\
& NMT & Nettoyez le filtre en papier. \cmark\\
& Google & Nettoyez le filtre à papier. \xmark\\
\\\multicolumn{3}{l}{\textbf{Common idioms}}\\
\hline
\multicolumn{3}{p{15cm}}{Is the English idiomatic expression correctly rendered with a suitable French idiomatic expression?}\\
\hline
S15a & Source & Stop \textbf{beating around the bush}. \\
& Ref & Cessez de \textbf{tourner autour du pot}. \\
& PBMT-1 & Cesser de battre la campagne. \xmark\\
& NMT & Arrêtez de battre autour de la brousse. \xmark\\
& Google & Arrêter de tourner autour du pot. \cmark\\
\end{tabular}
\end{table*}
\begin{table*}
\begin{tabular}{cll}
\hline
S15b & Source & You are \textbf{putting the cart before the horse}. \\
& Ref & Vous \textbf{mettez la charrue devant les bœufs}. \\
& PBMT-1 & Vous pouvez mettre la charrue avant les bœufs. \cmark\\
& NMT & Vous mettez la charrue avant le cheval. \xmark\\
& Google & Vous mettez le chariot devant le cheval. \xmark\\
\hline
S15c & Source & His comment proved to be \textbf{the straw that broke the camel's back}. \\
& Ref & Son commentaire s'est avéré être \textbf{la goutte d'eau qui a fait déborder le vase}. \\
& PBMT-1 & Son commentaire s'est révélé être la goutte d'eau qui fait déborder le vase. \cmark\\
& NMT & Son commentaire s'est avéré être la paille qui a brisé le dos du chameau. \xmark\\
& Google & Son commentaire s'est avéré être la paille qui a cassé le dos du chameau. \xmark\\
\hline
S15d & Source & His argument really \textbf{hit the nail on the head}. \\
& Ref & Son argument a vraiment \textbf{fait mouche}. \\
& PBMT-1 & Son argument a vraiment mis le doigt dessus. \cmark\\
& NMT & Son argument a vraiment frappé le clou sur la tête. \xmark\\
& Google & Son argument a vraiment frappé le clou sur la tête. \xmark\\
\hline
S15e & Source & It's \textbf{no use crying over spilt milk}. \\
& Ref & \textbf{Ce qui est fait est fait}. \\
& PBMT-1 & Ce n'est pas de pleurer sur le lait répandu. \xmark\\
& NMT & Il ne sert à rien de pleurer sur le lait haché. \xmark\\
& Google & Ce qui est fait est fait. \cmark\\
\hline
S15f & Source & It is \textbf{no use crying over spilt milk}. \\
& Ref & \textbf{Ce qui est fait est fait}. \\
& PBMT-1 & Il ne suffit pas de pleurer sur le lait répandu. \xmark\\
& NMT & Il ne sert à rien de pleurer sur le lait écrémé. \xmark\\
& Google & Il est inutile de pleurer sur le lait répandu. \xmark\\
\\\multicolumn{3}{l}{\textbf{Syntactically flexible idioms}}\\
\hline
\multicolumn{3}{p{15cm}}{Is the English idiomatic expression correctly rendered with a suitable French idiomatic expression?}\\
\hline
S16a & Source & The cart has been put before the horse. \\
& Ref & La \textbf{charrue a été mise devant les bœufs}. \\
& PBMT-1 & On met la charrue devant le cheval. \xmark\\
& NMT & Le chariot a été mis avant le cheval. \xmark\\
& Google & Le chariot a été mis devant le cheval. \xmark\\
\hline
S16b & Source & With this argument, \textbf{the nail has been hit on the head}. \\
& Ref & Avec cet argument, \textbf{la cause est entendue}. \\
& PBMT-1 & Avec cette argument, l'ongle a été frappée à la tête. \xmark\\
& NMT & Avec cet argument, l'ongle a été touché à la tête. \xmark\\
& Google & Avec cet argument, le clou a été frappé sur la tête. \xmark\\
\end{tabular}
\end{table*}
\begin{table*}
\begin{tabular}{cll}
\\\multicolumn{3}{l}{\large \textbf{Syntactic}}\\
\\\multicolumn{3}{l}{\textbf{Yes-no question syntax}}\\
\hline
\multicolumn{3}{p{15cm}}{Is the English question correctly rendered as a French question?}\\
\hline
S17a & Source & \textbf{Have the kids} ever watched that movie? \\
& Ref & \textbf{Les enfants ont-ils} déjà vu ce film? \\
& PBMT-1 & Les enfants jamais regardé ce film? \xmark\\
& NMT & Les enfants ont-ils déjà regardé ce film? \cmark\\
& Google & Les enfants ont-ils déjà regardé ce film? \cmark\\
\hline
S17b & Source & \textbf{Hasn't your boss denied you} a promotion? \\
& Ref & \textbf{Votre patron ne vous a-t-il pas refusé} une promotion? \\
& PBMT-1 & N'a pas nié votre patron vous un promotion? \xmark\\
& NMT & Est-ce que votre patron vous a refusé une promotion? \cmark\\
& Google & Votre patron ne vous a-t-il pas refusé une promotion? \cmark\\
\hline
S17c & Source & \textbf{Shouldn't I attend} this meeting? \\
& Ref & \textbf{Ne devrais-je pas assister} à cette réunion? \\
& PBMT-1 & Ne devrais-je pas assister à cette réunion? \cmark\\
& NMT & Est-ce que je ne devrais pas assister à cette réunion? \cmark\\
& Google & Ne devrais-je pas assister à cette réunion? \cmark\\
\\\multicolumn{3}{l}{\textbf{Tag questions}}\\
\hline
\multicolumn{3}{p{15cm}}{Is the English ``tag question'' element correctly rendered in the translation?}\\
\hline
S18a & Source & Mary looked really happy tonight, \textbf{didn't she}? \\
& Ref & Mary avait l'air vraiment heureuse ce soir, \textbf{n'est-ce pas}? \\
& PBMT-1 & Marie a regardé vraiment heureux de ce soir, n'est-ce pas elle? \xmark\\
& NMT & Mary s'est montrée vraiment heureuse ce soir, ne l'a pas fait? \xmark\\
& Google & Mary avait l'air vraiment heureuse ce soir, n'est-ce pas? \cmark\\
\hline
S18b & Source & We should not do that again, \textbf{should we}? \\
& Ref & Nous ne devrions pas refaire cela, \textbf{n'est-ce pas}? \\
& PBMT-1 & Nous ne devrions pas faire qu'une fois encore, faut-il? \xmark\\
& NMT & Nous ne devrions pas le faire encore, si nous? \xmark\\
& Google & Nous ne devrions pas recommencer, n'est-ce pas? \cmark\\
\hline
S18c & Source & She was perfect tonight, \textbf{was she not}? \\
& Ref & Elle était parfaite ce soir, \textbf{n'est-ce pas}? \\
& PBMT-1 & Elle était parfait ce soir, elle n'était pas? \xmark\\
& NMT & Elle était parfaite ce soir, n'était-elle pas? \xmark\\
& Google & Elle était parfaite ce soir, n'est-ce pas? \cmark\\
\\\multicolumn{3}{l}{\textbf{WH-MVT and stranded preps}}\\
\hline
\multicolumn{3}{p{15cm}}{Is the dangling preposition of the English sentence correctly placed in the French translation?}\\
\hline
S19a & Source & The guy \textbf{that} she is going out \textbf{with} is handsome. \\
& Ref & Le type \textbf{avec qui} elle sort est beau. \\
& PBMT-1 & Le mec qu'elle va sortir avec est beau. \xmark\\
& NMT & Le mec qu'elle sort avec est beau. \xmark\\
& Google & Le mec avec qui elle sort est beau. \cmark\\
\end{tabular}
\end{table*}
\begin{table*}
\begin{tabular}{cll}
\hline
S19b & Source & \textbf{Whom} is she going out \textbf{with} these days? \\
& Ref & \textbf{Avec qui} sort-elle ces jours-ci? \\
& PBMT-1 & Qu'est-ce qu'elle allait sortir avec ces jours? \xmark\\
& NMT & À qui s'adresse ces jours-ci? \xmark\\
& Google & Avec qui sort-elle de nos jours? \cmark\\
\hline
S19c & Source & The girl \textbf{that} he has been talking \textbf{about} is smart. \\
& Ref & La fille \textbf{dont} il a parlé est brillante. \\
& PBMT-1 & La jeune fille qu'il a parlé est intelligent. \xmark\\
& NMT & La fille qu'il a parlé est intelligente. \xmark\\
& Google & La fille dont il a parlé est intelligente. \cmark\\
\hline
S19d & Source & \textbf{Who} was he talking \textbf{to} when you left? \\
& Ref & \textbf{À qui} parlait-il au moment où tu es parti? \\
& PBMT-1 & Qui est lui parler quand vous avez quitté? \xmark\\
& NMT & Qui a-t-il parlé à quand vous avez quitté? \xmark\\
& Google & Avec qui il parlait quand vous êtes parti? \cmark\\
\hline
S19e & Source & The city \textbf{that} he is arriving \textbf{from} is dangerous. \\
& Ref & La ville \textbf{d'où} il arrive est dangereuse. \\
& PBMT-1 & La ville qu'il est arrivé de est dangereuse. \xmark\\
& NMT & La ville qu'il est en train d'arriver est dangereuse. \xmark\\
& Google & La ville d'où il vient est dangereuse. \cmark\\
\hline
S19f & Source & \textbf{Where} is he arriving \textbf{from}? \\
& Ref & \textbf{D'où} arrive-t-il? \\
& PBMT-1 & Où est-il arrivé? \xmark\\
& NMT & De quoi s'agit-il? \xmark\\
& Google & D'où vient-il? \cmark\\
\\\multicolumn{3}{l}{\textbf{Adverb-triggered inversion}}\\
\hline
\multicolumn{3}{p{15cm}}{Is the adverb-triggered subject-verb inversion in the English sentence correctly rendered in the French translation?}\\
\hline
S20a & Source & Rarely \textbf{did the dog} run. \\
& Ref & Rarement \textbf{le chien courait-il}. \\
& PBMT-1 & Rarement le chien courir. \xmark\\
& NMT & Il est rare que le chien marche. \xmark\\
& Google & Rarement le chien courir. \xmark\\
\hline
S20b & Source & Never before \textbf{had she been} so unhappy. \\
& Ref & Jamais encore \textbf{n'avait-elle} été aussi malheureuse. \\
& PBMT-1 & Jamais auparavant, si elle avait été si malheureux. \xmark\\
& NMT & Jamais auparavant n'avait été si malheureuse. \xmark\\
& Google & Jamais elle n'avait été aussi malheureuse. \cmark\\
\end{tabular}
\end{table*}
\begin{table*}
\begin{tabular}{cll}
\hline
S20c & Source & Nowhere \textbf{were the birds} so colorful. \\
& Ref & Nulle part \textbf{les oiseaux} \textbf{n'étaient} si colorés. \\
& PBMT-1 & Nulle part les oiseaux de façon colorée. \xmark\\
& NMT & Les oiseaux ne sont pas si colorés. \xmark\\
& Google & Nulle part les oiseaux étaient si colorés. \xmark\\
\\\multicolumn{3}{l}{\textbf{Middle voice}}\\
\hline
\multicolumn{3}{p{15cm}}{Is the generic statement made in the English sentence correctly and naturally rendered in the French translation? }\\
\hline
S21a & Source & Soup \textbf{is eaten} with a large spoon. \\
& Ref & La soupe \textbf{se mange} avec une grande cuillère \\
& PBMT-1 & La soupe est mangé avec une grande cuillère. \xmark\\
& NMT & La soupe est consommée avec une grosse cuillère. \xmark\\
& Google & La soupe est consommée avec une grande cuillère. \xmark\\
\hline
S21b & Source & Masonry \textbf{is cut} using a diamond blade. \\
& Ref & La maçonnerie \textbf{se coupe} avec une lame à diamant. \\
& PBMT-1 & La maçonnerie est coupé à l'aide d'une lame de diamant. \xmark\\
& NMT & La maçonnerie est coupée à l'aide d'une lame de diamant. \xmark\\
& Google & La maçonnerie est coupée à l'aide d'une lame de diamant. \xmark\\
\hline
S21c & Source & Champagne \textbf{is drunk} in a glass called a flute. \\
& Ref & Le champagne \textbf{se boit} dans un verre appelé flûte. \\
& PBMT-1 & Le champagne est ivre dans un verre appelé une flûte. \xmark\\
& NMT & Le champagne est ivre dans un verre appelé flûte. \xmark\\
& Google & Le Champagne est bu dans un verre appelé flûte. \xmark\\
\\\multicolumn{3}{l}{\textbf{Fronted ``should''}}\\
\hline
\multicolumn{3}{p{15cm}}{Fronted ``should'' is interpreted as a conditional subordinator. It is normally translated as ``si'' with imperfect tense.}\\
\hline
S22a & Source & \textbf{Should} Paul leave, I would be sad. \\
& Ref & \textbf{Si} Paul \textbf{devait} s'en aller, je serais triste. \\
& PBMT-1 & Si le congé de Paul, je serais triste. \xmark\\
& NMT & Si Paul quitte, je serais triste. \xmark\\
& Google & Si Paul s'en allait, je serais triste. \cmark\\
\hline
S22b & Source & Should he become president, she would be promoted immediately. \\
& Ref & \textbf{S'}il devait devenir président, elle recevrait immédiatement une promotion. \\
& PBMT-1 & S'il devait devenir président, elle serait encouragée immédiatement. \cmark\\
& NMT & S'il devait devenir président, elle serait immédiatement promue. \cmark\\
& Google & Devrait-il devenir président, elle serait immédiatement promue. \xmark\\
\hline
S22c & Source & \textbf{Should} he fall, he would get up again immediately. \\
& Ref & \textbf{S'} il venait à tomber, il se relèverait immédiatement. \\
& PBMT-1 & S'il devait tomber, il allait se lever immédiatement de nouveau. \cmark\\
& NMT & S'il tombe, il serait de nouveau immédiatement. \xmark\\
& Google & S'il tombe, il se lèvera immédiatement. \xmark\\
\end{tabular}
\end{table*}
\begin{table*}
\begin{tabular}{cll}
\\\multicolumn{3}{l}{\textbf{Clitic pronouns}}\\
\hline
\multicolumn{3}{p{15cm}}{Are the English pronouns correctly rendered in the French translations?}\\
\hline
S23a & Source & She had a lot of money but he did not have \textbf{any}. \\
& Ref & Elle avait beaucoup d'argent mais il n'\textbf{en} avait pas. \\
& PBMT-1 & Elle avait beaucoup d'argent mais il n'en avait pas. \cmark\\
& NMT & Elle avait beaucoup d'argent, mais il n'a pas eu d'argent. \cmark\\
& Google & Elle avait beaucoup d'argent mais il n'en avait pas. \cmark\\
\hline
S23b & Source & He did not talk \textbf{to them} very often. \\
& Ref & Il ne \textbf{leur} parlait pas très souvent. \\
& PBMT-1 & Il n'a pas leur parler très souvent. \xmark\\
& NMT & Il ne leur a pas parlé très souvent. \cmark\\
& Google & Il ne leur parlait pas très souvent. \cmark\\
\hline
S23c & Source & The men are watching \textbf{each other}. \\
& Ref & Les hommes \textbf{se} surveillent \textbf{l'un l'autre} \\
& PBMT-1 & Les hommes se regardent les uns les autres. \cmark\\
& NMT & Les hommes se regardent les uns les autres. \cmark\\
& Google & Les hommes se regardent. \xmark\\
\hline
S23d & Source & He gave \textbf{it} to the man. \\
& Ref & Il \textbf{le} donna à l'homme. \\
& PBMT-1 & Il a donné à l'homme. \xmark\\
& NMT & Il l'a donné à l'homme. \cmark\\
& Google & Il le donna à l'homme. \cmark\\
\hline
S23e & Source & He did not give \textbf{it} to \textbf{her}. \\
& Ref & Il ne \textbf{le} \textbf{lui} a pas donné. \\
& PBMT-1 & Il ne lui donner. \xmark\\
& NMT & Il ne l'a pas donné à elle. \xmark\\
& Google & Il ne lui a pas donné. \xmark\\
\\\multicolumn{3}{l}{\textbf{Ordinal placement}}\\
\hline
\multicolumn{3}{p{15cm}}{Is the relative order of the ordinals and numerals correct in the French tranlation?}\\
\hline
S24a & Source & The \textbf{first four} men were exhausted. \\
& Ref & Les \textbf{quatre premiers} hommes étaient tous épuisés. \\
& PBMT-1 & Les quatre premiers hommes étaient épuisés. \cmark\\
& NMT & Les quatre premiers hommes ont été épuisés. \cmark\\
& Google & Les quatre premiers hommes étaient épuisés. \cmark\\
\hline
S24b & Source & The \textbf{last three} candidates were eliminated. \\
& Ref & Les \textbf{trois derniers} candidats ont été éliminés. \\
& PBMT-1 & Les trois derniers candidats ont été éliminés. \cmark\\
& NMT & Les trois derniers candidats ont été éliminés. \cmark\\
& Google & Les trois derniers candidats ont été éliminés. \cmark\\
\end{tabular}
\end{table*}
\begin{table*}
\begin{tabular}{cll}
\hline
S24c & Source & The \textbf{other two} guys left without paying. \\
& Ref & Les \textbf{deux autres} types sont partis sans payer. \\
& PBMT-1 & Les deux autres mecs ont laissé sans payer. \cmark\\
& NMT & Les deux autres gars à gauche sans payer. \cmark\\
& Google & Les deux autres gars sont partis sans payer. \cmark\\
\\\multicolumn{3}{l}{\textbf{Inalienable possession}}\\
\hline
\multicolumn{3}{p{15cm}}{Is the French translation correct and natural both in: a) its use of a particular determiner on the body part noun; and b) the presence or absence of a reflexive pronoun before the verb?}\\
\hline
S25a & Source & He washed \textbf{his} hands. \\
& Ref & Il \textbf{s'}est lavé \textbf{les} mains. \\
& PBMT-1 & Il se lavait les mains. \cmark\\
& NMT & Il a lavé ses mains. \xmark\\
& Google & Il se lava les mains. \cmark\\
\hline
S25b & Source & I brushed \textbf{my} teeth. \\
& Ref & Je \textbf{me} suis brossé \textbf{les} dents. \\
& PBMT-1 & J'ai brossé mes dents. \xmark\\
& NMT & J'ai brossé mes dents. \xmark\\
& Google & Je me suis brossé les dents. \cmark\\
\hline
S25c & Source & You brushed \textbf{your} teeth. \\
& Ref & Tu \textbf{t'}es brossé \textbf{les} dents \\
& PBMT-1 & Vous avez brossé vos dents. \xmark\\
& NMT & vous avez brossé vos dents. \xmark\\
& Google & Tu as brossé les dents. \xmark\\
\hline
S25d & Source & I raised \textbf{my} hand. \\
& Ref & J'ai levé \textbf{la} main. \\
& PBMT-1 & J'ai levé la main. \cmark\\
& NMT & J'ai soulevé ma main. \xmark\\
& Google & Je levai la main. \cmark\\
\hline
S25e & Source & He turned \textbf{his} head. \\
& Ref & Il a tourné \textbf{la} tête. \\
& PBMT-1 & Il a transformé sa tête. \xmark\\
& NMT & Il a tourné sa tête. \xmark\\
& Google & Il tourna la tête. \cmark\\
\hline
S25f & Source & He raised his eyes to heaven. \\
& Ref & Il leva \textbf{les} yeux au ciel. \\
& PBMT-1 & Il a évoqué les yeux au ciel. \cmark\\
& NMT & Il a levé les yeux sur le ciel. \cmark\\
& Google & Il leva les yeux au ciel. \cmark\\
\end{tabular}
\end{table*}
\begin{table*}
\begin{tabular}{cll}
\\\multicolumn{3}{l}{\textbf{Zero REL PRO}}\\
\hline
\multicolumn{3}{p{15cm}}{Is the English zero relative pronoun correctly translated as a non-zero one in the French translation?}\\
\hline
S26a & Source & The strangers \textbf{} the woman saw were working. \\
& Ref & Les inconnus \textbf{que} la femme vit travaillaient. \\
& PBMT-1 & Les étrangers la femme vit travaillaient. \xmark\\
& NMT & Les inconnus de la femme ont travaillé. \xmark\\
& Google & Les étrangers que la femme vit travaillaient. \cmark\\
\hline
S26b & Source & The man \textbf{} your sister hates is evil. \\
& Ref & L'homme \textbf{que} votre sœur déteste est méchant. \\
& PBMT-1 & L'homme ta soeur hait est le mal. \xmark\\
& NMT & L'homme que ta soeur est le mal est le mal. \cmark\\
& Google & L'homme que votre sœur hait est méchant. \cmark\\
\hline
S26c & Source & The girl \textbf{} my friend was talking about is gone. \\
& Ref & La fille \textbf{dont} mon ami parlait est partie. \\
& PBMT-1 & La jeune fille mon ami a parlé a disparu. \xmark\\
& NMT & La petite fille de mon ami était révolue. \xmark\\
& Google & La fille dont mon ami parlait est partie. \cmark\\
\end{tabular}
\end{table*}

%% file: challenge_eval.bbl
\begin{thebibliography}{}
\expandafter\ifx\csname natexlab\endcsname\relax\def\natexlab#1{#1}\fi

\bibitem[{Bahdanau et~al.(2015)Bahdanau, Cho, and Bengio}]{Bahdanau15}
Dzmitry Bahdanau, Kyunghyun Cho, and Yoshua Bengio. 2015.
\newblock {Neural machine translation by
  jointly learning to align and translate}.
\newblock In {\em Proceedings of the Third International Conference on Learning
  Representations (ICLR)\/}. San Diego, USA.
\newblock
  \href{http://arxiv.org/abs/1409.0473}{http://arxiv.org/abs/1409.0473}.

\bibitem[{Bentivogli et~al.(2016)Bentivogli, Bisazza, Cettolo, and
  Federico}]{Bentivogli:EMNLP2016}
Luisa Bentivogli, Arianna Bisazza, Mauro Cettolo, and Marcello Federico. 2016.
\newblock {Neural versus
  phrase-based machine translation quality: a case study}.
\newblock In {\em Proceedings of the 2016 Conference on Empirical Methods in
  Natural Language Processing\/}. Association for Computational Linguistics,
  Austin, Texas, pages 257--267.
\newblock
  \href{https://aclweb.org/anthology/D16-1025}{https://aclweb.org/anthology/D16-1025}.

\bibitem[{Bojar et~al.(2016)Bojar, Chatterjee, Federmann, Graham, Haddow, Huck,
  Jimeno~Yepes, Koehn, Logacheva, Monz, Negri, Neveol, Neves, Popel, Post,
  Rubino, Scarton, Specia, Turchi, Verspoor, and Zampieri}]{Bojar:WMT2016}
Ond\v{r}ej Bojar, Rajen Chatterjee, Christian Federmann, Yvette Graham, Barry
  Haddow, Matthias Huck, Antonio Jimeno~Yepes, Philipp Koehn, Varvara
  Logacheva, Christof Monz, Matteo Negri, Aurelie Neveol, Mariana Neves, Martin
  Popel, Matt Post, Raphael Rubino, Carolina Scarton, Lucia Specia, Marco
  Turchi, Karin Verspoor, and Marcos Zampieri. 2016.
\newblock {Findings of the 2016
  conference on machine translation}.
\newblock In {\em Proceedings of the First Conference on Machine
  Translation\/}. Association for Computational Linguistics, Berlin, Germany,
  pages 131--198.
\newblock
  \href{http://www.aclweb.org/anthology/W16-2301}{http://www.aclweb.org/anthology/W16-2301}.

\bibitem[{Cherry(2013)}]{Cherry2013}
Colin Cherry. 2013.
\newblock {Improved reordering
  for phrase-based translation using sparse features}.
\newblock In {\em Proceedings of the 2013 Conference of the North American
  Chapter of the Association for Computational Linguistics: Human Language
  Technologies\/}. Association for Computational Linguistics, Atlanta, Georgia,
  pages 22--31.
\newblock
  \href{http://www.aclweb.org/anthology/N13-1003}{http://www.aclweb.org/anthology/N13-1003}.

\bibitem[{Cherry and Foster(2012)}]{CherryFoster2012}
Colin Cherry and George Foster. 2012.
\newblock {Batch tuning
  strategies for statistical machine translation}.
\newblock In {\em Proceedings of the 2012 Conference of the North American
  Chapter of the Association for Computational Linguistics: Human Language
  Technologies\/}. Association for Computational Linguistics, Montr\'{e}al,
  Canada, pages 427--436.
\newblock
  \href{http://www.aclweb.org/anthology/N12-1047}{http://www.aclweb.org/anthology/N12-1047}.

\bibitem[{Cho et~al.(2014)Cho, van Merrienboer, Gulcehre, Bahdanau, Bougares,
  Schwenk, and Bengio}]{Cho14}
Kyunghyun Cho, Bart van Merrienboer, Caglar Gulcehre, Dzmitry Bahdanau, Fethi
  Bougares, Holger Schwenk, and Yoshua Bengio. 2014.
\newblock {Learning phrase
  representations using rnn encoder--decoder for statistical machine
  translation}.
\newblock In {\em Proceedings of the 2014 Conference on Empirical Methods in
  Natural Language Processing (EMNLP)\/}. Association for Computational
  Linguistics, Doha, Qatar, pages 1724--1734.
\newblock
  \href{http://www.aclweb.org/anthology/D14-1179}{http://www.aclweb.org/anthology/D14-1179}.

\bibitem[{Devlin et~al.(2014)Devlin, Zbib, Huang, Lamar, Schwartz, and
  Makhoul}]{Devlin:14}
Jacob Devlin, Rabih Zbib, Zhongqiang Huang, Thomas Lamar, Richard Schwartz, and
  John Makhoul. 2014.
\newblock {Fast and robust
  neural network joint models for statistical machine translation}.
\newblock In {\em Proceedings of the 52nd Annual Meeting of the Association for
  Computational Linguistics (Volume 1: Long Papers)\/}. Association for
  Computational Linguistics, Baltimore, Maryland, pages 1370--1380.
\newblock
  \href{http://www.aclweb.org/anthology/P14-1129}{http://www.aclweb.org/anthology/P14-1129}.

\bibitem[{Dorr(1994)}]{Dorr:94}
Bonnie~J. Dorr. 1994.
\newblock {Machine
  translation divergences: a formal description and proposed solution}.
\newblock {\em Computational Linguistics\/} 20:4.
\newblock
  \href{http://aclweb.org/anthology/J/J94/J94-4004.pdf}{http://aclweb.org/anthology/J/J94/J94-4004.pdf}.

\bibitem[{Durrani et~al.(2016)Durrani, Dalvi, Sajjad, and
  Vogel}]{Durrani:IWSLT2016}
Nadir Durrani, Fahim Dalvi, Hassan Sajjad, and Stephan Vogel. 2016.
\newblock
  \href{https://workshop2016.iwslt.org/downloads/qcri-machine-translation.pdf}{{QCRI}
  machine translation systems for {IWSLT} 16}.
\newblock In {\em Proceedings of the 13th International Workshop on Spoken
  Language Translation (IWSLT)\/}. Seattle, Washington.

\bibitem[{Galley and Manning(2008)}]{Galley08}
Michel Galley and Christopher~D. Manning. 2008.
\newblock {A simple and
  effective hierarchical phrase reordering model}.
\newblock In {\em Proceedings of the 2008 Conference on Empirical Methods in
  Natural Language Processing\/}. Association for Computational Linguistics,
  Honolulu, Hawaii, pages 848--856.
\newblock
  \href{http://www.aclweb.org/anthology/D08-1089}{http://www.aclweb.org/anthology/D08-1089}.

\bibitem[{Huang and Chiang(2007)}]{HuangChiang2007}
Liang Huang and David Chiang. 2007.
\newblock {Forest rescoring:
  Faster decoding with integrated language models}.
\newblock In {\em Proceedings of the 45th Annual Meeting of the Association of
  Computational Linguistics\/}. Association for Computational Linguistics,
  Prague, Czech Republic, pages 144--151.
\newblock
  \href{http://www.aclweb.org/anthology/P07-1019}{http://www.aclweb.org/anthology/P07-1019}.

\bibitem[{Junczys-Dowmunt et~al.(2016{\natexlab{a}})Junczys-Dowmunt, Dwojak,
  and Hoang}]{Junczys-Dowmunt:IWSLT2016}
Marcin Junczys-Dowmunt, Tomasz Dwojak, and Hieu Hoang. 2016{\natexlab{a}}.
\newblock Is neural machine translation ready for deployment? a case study on
  30 translation directions.
\newblock In {\em Proceedings of the 13th International Workshop on Spoken
  Language Translation (IWSLT)\/}. Seattle, Washington.

\bibitem[{Junczys-Dowmunt et~al.(2016{\natexlab{b}})Junczys-Dowmunt, Dwojak,
  and Sennrich}]{Junczys-Dowmunt16}
Marcin Junczys-Dowmunt, Tomasz Dwojak, and Rico Sennrich. 2016{\natexlab{b}}.
\newblock {The amu-uedin
  submission to the wmt16 news translation task: Attention-based nmt models as
  feature functions in phrase-based smt}.
\newblock In {\em Proceedings of the First Conference on Machine
  Translation\/}. Association for Computational Linguistics, Berlin, Germany,
  pages 319--325.
\newblock
  \href{http://www.aclweb.org/anthology/W16-2316}{http://www.aclweb.org/anthology/W16-2316}.

\bibitem[{Kalchbrenner and Blunsom(2013)}]{Kalchbrenner13}
Nal Kalchbrenner and Phil Blunsom. 2013.
\newblock {Recurrent continuous
  translation models}.
\newblock In {\em Proceedings of the 2013 Conference on Empirical Methods in
  Natural Language Processing\/}. Association for Computational Linguistics,
  Seattle, Washington, USA, pages 1700--1709.
\newblock
  \href{http://www.aclweb.org/anthology/D13-1176}{http://www.aclweb.org/anthology/D13-1176}.

\bibitem[{King and Falkedal(1990)}]{King90}
Margaret King and Kirsten Falkedal. 1990.
\newblock {Using test
  suites in evaluation of machine translation systems}.
\newblock In {\em Proceedings of the 1990 Conference on Computational
  Linguistics\/}. Association for Computational Linguistics, Helsinki, Finland.
\newblock
  \href{http://aclweb.org/anthology/C/C90/C90-2037.pdf}{http://aclweb.org/anthology/C/C90/C90-2037.pdf}.

\bibitem[{Papineni et~al.(2002)Papineni, Roukos, Ward, and
  Zhu}]{Papineni:ACL2002}
Kishore Papineni, Salim Roukos, Todd Ward, and Wei-Jing Zhu. 2002.
\newblock {Bleu: a method for
  automatic evaluation of machine translation}.
\newblock In {\em Proceedings of 40th Annual Meeting of the Association for
  Computational Linguistics\/}. Association for Computational Linguistics,
  Philadelphia, Pennsylvania, USA, pages 311--318.
\newblock
  \href{https://doi.org/10.3115/1073083.1073135}{https://doi.org/10.3115/1073083.1073135}.

\bibitem[{Sennrich(2016)}]{Sennrich:arXiv2016}
Rico Sennrich. 2016.
\newblock {How grammatical is
  character-level neural machine translation? assessing {MT} quality with
  contrastive translation pairs}.
\newblock {\em CoRR\/} abs/1612.04629.
\newblock
  \href{http://arxiv.org/abs/1612.04629}{http://arxiv.org/abs/1612.04629}.

\bibitem[{Sennrich et~al.(2016)Sennrich, Haddow, and Birch}]{Sennrich15}
Rico Sennrich, Barry Haddow, and Alexandra Birch. 2016.
\newblock {Neural machine
  translation of rare words with subword units}.
\newblock In {\em Proceedings of the 54th Annual Meeting of the Association for
  Computational Linguistics (Volume 1: Long Papers)\/}. Association for
  Computational Linguistics, Berlin, Germany, pages 1715--1725.
\newblock
  \href{http://www.aclweb.org/anthology/P16-1162}{http://www.aclweb.org/anthology/P16-1162}.

\bibitem[{Sutskever et~al.(2014)Sutskever, Vinyals, and Le}]{Sutskever14}
Ilya Sutskever, Oriol Vinyals, and Quoc~V Le. 2014.
\newblock
  \href{http://papers.nips.cc/paper/5346-sequence-to-sequence-learning-with-neural-networks.pdf}{Sequence
  to sequence learning with neural networks}.
\newblock In {\em Advances in Neural Information Processing Systems 27\/}.
  Curran Associates, Inc., pages 3104--3112.

\bibitem[{Toral and Sánchez-Cartagena(2017)}]{Toral:EACL2017}
Antonio Toral and Víctor~M. Sánchez-Cartagena. 2017.
\newblock {A multifaceted
  evaluation of neural versus statistical machine translation for 9 language
  directions}.
\newblock In {\em Proceedings of the The 15th Conference of the European
  Chapter of the Association for Computational Linguistics (EACL)\/}. Valencia,
  Spain, pages 1063--1073.
\newblock
  \href{http://aclweb.org/anthology/E/E17/E17-1100.pdf}{http://aclweb.org/anthology/E/E17/E17-1100.pdf}.

\bibitem[{Vinay and Darbelnet(1958)}]{Vinay:58}
Jean-Paul Vinay and Jean Darbelnet. 1958.
\newblock {\em Stylistique comparée du français et de l'anglais\/}, volume~1.
\newblock Didier, Paris.

\bibitem[{Wu et~al.(2016)Wu, Schuster, Chen, Le, Norouzi, Macherey, Krikun,
  Cao, Gao, Macherey, Klingner, Shah, Johnson, Liu, Kaiser, Gouws, Kato, Kudo,
  Kazawa, Stevens, Kurian, Patil, Wang, Young, Smith, Riesa, Rudnick, Vinyals,
  Corrado, Hughes, and Dean}]{Wu16}
Yonghui Wu, Mike Schuster, Zhifeng Chen, Quoc~V. Le, Mohammad Norouzi, Wolfgang
  Macherey, Maxim Krikun, Yuan Cao, Qin Gao, Klaus Macherey, Jeff Klingner,
  Apurva Shah, Melvin Johnson, Xiaobing Liu, Lukasz Kaiser, Stephan Gouws,
  Yoshikiyo Kato, Taku Kudo, Hideto Kazawa, Keith Stevens, George Kurian,
  Nishant Patil, Wei Wang, Cliff Young, Jason Smith, Jason Riesa, Alex Rudnick,
  Oriol Vinyals, Greg Corrado, Macduff Hughes, and Jeffrey Dean. 2016.
\newblock {Google's neural machine
  translation system: Bridging the gap between human and machine translation}.
\newblock {\em CoRR\/} abs/1609.08144.
\newblock
  \href{http://arxiv.org/abs/1609.08144}{http://arxiv.org/abs/1609.08144}.

\bibitem[{Zeiler(2012)}]{Zeiler12}
Matthew~D. Zeiler. 2012.
\newblock {{ADADELTA:} an adaptive
  learning rate method}.
\newblock {\em CoRR\/} abs/1212.5701.
\newblock
  \href{http://arxiv.org/abs/1212.5701}{http://arxiv.org/abs/1212.5701}.

\end{thebibliography}
